\documentclass[10pt,twocolumn,letterpaper]{article}

\usepackage{iccv}              %

\usepackage{tabularx} %
\usepackage{xspace} %
\usepackage{xcolor} %
\usepackage{makecell} %
\usepackage{multirow} %
\usepackage{bm} %
\usepackage{mathtools}  %
\usepackage{colortbl} %
\usepackage{subcaption}  %
\usepackage[symbol]{footmisc}  %
\usepackage[export]{adjustbox}

\newcolumntype{Y}{>{\centering\arraybackslash}X}
\newcolumntype{P}[1]{>{\centering\arraybackslash}p{#1}}
\newcolumntype{C}{>{\centering}X}

\newcommand{\OURS}{Avat3r}

\definecolor{iccvblue}{rgb}{0.21,0.49,0.74}
\usepackage[pagebackref,breaklinks,colorlinks,allcolors=iccvblue]{hyperref}

\def\paperID{8454} %
\def\confName{ICCV}
\def\confYear{2025}

\title{Avat3r: Large Animatable Gaussian Reconstruction Model \\for High-fidelity 3D Head Avatars}

\author{Tobias Kirschstein$^{1, 2^*}$ \, Javier Romero$^{2}$ \, Artem Sevastopolsky$^{1, 2^*}$ \, Matthias Nießner$^{1}$ \, Shunsuke Saito$^{2}$
  \\
  Technical University of Munich$^{1}$ \qquad Meta Reality Labs Pittsburgh$^{2}$}

\begin{document}
\twocolumn[{
\renewcommand\twocolumn[1][]{#1}%
\maketitle
\vspace{-0.5cm}
\includegraphics[width=\textwidth]{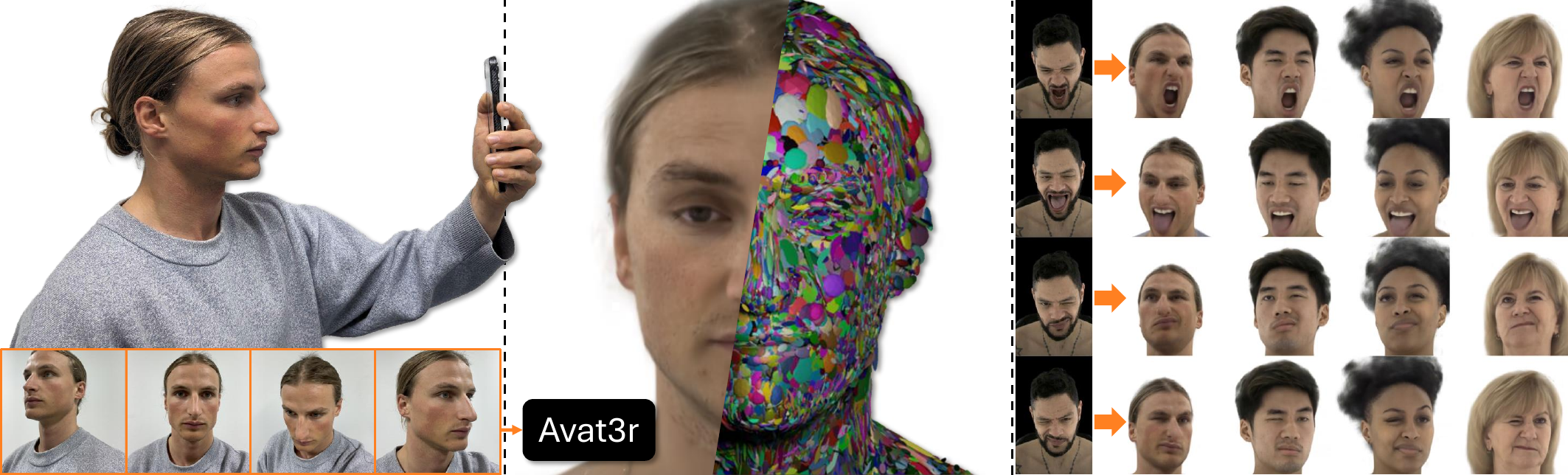}
\begin{tabularx}{\linewidth}{P{0.3\linewidth}P{0.29\linewidth}P{0.35\linewidth}}
(a) Phone scan & (b) Sparse 3D head reconstruction & (c) Zero-shot 3D facial animation
\end{tabularx}
\captionof{figure}{\textbf{Avat3r.} Given just four images of a person's head (a), {\OURS} achieves two things: (b) it creates a faithful 3D reconstruction of the head in a feed-forward manner, and (c) it allows facial animation without having seen any of the corresponding expressions of the person. This simplifies the capturing process as it drops the need for recording long sequences of facial movement. As a result, the entire pipeline from head scan to final 3D head avatar can be executed within a few minutes and runs on a single consumer-grade GPU.} \vspace{2em}
\label{fig:teaser}
}]

\maketitle

\let\thefootnote\relax\footnotetext{$^*$Work done while Tobias Kirschstein and Artem Sevastopolsky were interns at Reality Labs Research, Pittsburgh, PA, USA}\par
    \let\thefootnote\relax\footnotetext{
    \urlstyle{same}
Project website: {\url{https://tobias-kirschstein.github.io/avat3r/}}}\par
\begin{abstract}
Traditionally, creating photo-realistic 3D head avatars requires a studio-level multi-view capture setup and expensive optimization during test-time, limiting the use of digital human doubles to the VFX industry or offline renderings. 
To address this shortcoming, we present {\OURS}, which regresses a high-quality and animatable 3D head avatar from just a few input images, vastly reducing compute requirements during inference. 
More specifically, we make Large Reconstruction Models animatable and learn a powerful prior over 3D human heads from a large multi-view video dataset. %
For better 3D head reconstructions, we employ position maps from DUSt3R and generalized feature maps from the human foundation model Sapiens. %
To animate the 3D head, our key discovery is that simple cross-attention to an expression code is already sufficient.
Finally, we increase robustness by feeding input images with different expressions to our model during training, enabling the reconstruction of 3D head avatars from inconsistent inputs, e.g., an imperfect phone capture with accidental movement, or frames from a monocular video. 
We compare {\OURS} with current state-of-the-art methods for few-input and single-input scenarios, and find that our method has a competitive advantage in both tasks.   
Finally, we demonstrate the wide applicability of our proposed model, creating 3D head avatars from images of different sources, smartphone captures, single images, and even out-of-domain inputs like antique busts.
\end{abstract}

\section{Introduction}

Lifelike 3D head avatars find use in a variety of applications such as telepresence, movie production, personalized games, or digital education. Typically, the creation of a high-fidelity avatar requires studio-quality videos that have to be recorded in hour-long sessions in expensive capture setups. This limits the applicability of a 3D head avatar creation pipeline, especially in casual settings where users may only want to use a few images taken with their smartphone to create a convincing digital double. In this paper, we present Avat3r, a method that can reconstruct an animatable 3D head avatar from just a few images of a person. 

Building such a system involves several hard challenges. First, sparse 3D reconstruction poses a heavily under-constrained optimization problem since some regions such as the mouth interior or the side of the head may have simply not been seen in the available images. Next, animating the head requires plausibly morphing the reconstructed face without ever having seen that particular person talking or sticking their tongue out. Finally, the input images may be inconsistent, e.g., because the person could not remain completely static during a casual capture.
All of these three challenges, i.e., sparse 3D reconstruction, face animation, and robust reconstruction, have to be solved simultaneously to successfully reconstruct a photo-realistic 3D head avatar given just a few images.

Due to the aforementioned challenges, the current best methods for high-quality avatar creation~\cite{giebenhain2024npga, saito2024rgca, qian2024gaussianavatars, kirschstein2024diffusionavatars, teotia2024gaussianheads, teotia2024hq3davatar, kabadayi2024ganavatar, zielonka2023d3ga, xu2023gha} cannot be applied since their optimization-based approach requires many more observations than just a few images of a person. 
On the other hand, methods that aim at 3D-aware portrait animation~\cite{drobyshev2022megaportraits, tran2024voodoo3d, chu2024gagavatar, chu2024gpavatar} learn a prior over 3D heads from large video datasets.
As a result, existing methods can already convincingly animate faces from as few as just a single input image. However, such methods typically focus on frontal facial appearance and may sacrifice 3D-consistency or temporal consistency to reach satisfying image synthesis quality. This makes these methods less applicable in scenarios where a proper 3D model is required.
Another approach is to build a photo-realistic 3D head model that learns the distribution of 3D face geometry, expression, and appearance, akin to classic 3DMMs. While such unconditional generative models are very versatile and recent works have shown promising results~\cite{xu2025gphm, zheng2024headgap}, they typically suffer from limited generalizability along the identity axis due to existing 3D face datasets only providing a few hundred different persons~\cite{kirschstein2023nersemble, ava256, yang2020facescape, pan2024renderme360}. 

Our approach is motivated by the observation that 3D data for faces is limited in the identity axis, but not in the expression axis that is already well captured in existing datasets. We therefore design a system that is \textit{conditional} on the identity of a person, but \textit{generalizes} along the expression axis. Opposed to photo-realistic 3D face models, the network does not have to learn the full extent of human facial appearance but just needs to understand how to reconstruct it from a few example images, which simplifies the task. Inspired by recent Large Reconstruction Models~\cite{tang2025lgm, xu2024grm, zhang2025gslrm}, we devise an architecture that predicts 3D Gaussians for each pixel in the input images. As such, we deliberately avoid anchoring the 3D Gaussians on a template mesh, as common in the 3D Head Avatar domain~\cite{qian2024gaussianavatars, saito2024rgca, cao2022authentic}. The reason is that human facial appearance is complex and diverse. Limiting the 3D representation to a fixed number of Gaussians anchored on a template mesh with uniform topology will cause issues in regions such as hair or face accessories. Instead, when predicting Gaussians for each foreground pixel, a person with voluminous hair will receive more primitives to model their head than a bald person.
To simplify the task of sparse 3D reconstruction, we enrich each input image with feature maps from Sapiens~\cite{khirodkar2025sapiens} and position maps from DUSt3R~\cite{wang2024dust3r}. These position maps are injected into the network via skip connections and act as a coarse starting position for each Gaussian. We find that DUSt3R is still capable of producing reasonable position maps even if the input images are not consistent.
We exploit this remarkable capability of DUSt3R to train our model on input images taken at different timesteps, which not only enables training and inference on larger monocular video datasets but also makes the model more robust to inconsistencies in the input images. Such inconsistencies can quickly occur in a real-life scenario where a person may not stay completely still during a phone capture of their head.  Finally, to model the animation of the face, we find that simple cross-attention between inter\-mediate feature maps and a descriptive expression code are already sufficient to generalize over the space of facial expressions.
Taken together, our contributions are as follows:
\begin{itemize}
    \item We design Avat3r, a novel pipeline for creating high-quality 3D head avatars from just a few images in a matter of minutes.
    \item We build the first animatable large 3D reconstruction model and show that simple cross-attention on expression codes is sufficient to model complex facial animations.
    \item To improve robustness, we integrate priors from Dust3r and Sapiens via skip connections, and feed input images from different timesteps to the network during training.
\end{itemize}
 
\section{Related Work}

\subsection{Large 3D Reconstruction Models}
Recently, generalized feed-forward models for 3D reconstruction from sparse input views have garnered considerable attention due to their applicability in heavily under-constrained scenarios. The Large Reconstruction Model (LRM)~\cite{hong2023lrm} uses a transformer-based encoder-decoder pipeline to infer a NeRF reconstruction from just a single image. Newer iterations have shifted the focus towards generating 3D Gaussian representations from four input images~\cite{tang2025lgm, xu2024grm, zhang2025gslrm, charatan2024pixelsplat, chen2025mvsplat, liu2025mvsgaussian}, showing remarkable novel view synthesis results. Very recently, FaceLift~\cite{lyu2024facelift} has proposed a combination of personalized multi-view image diffusion and a large reconstruction model for single-image 3D head reconstruction. The paradigm of transformer-based sparse 3D reconstruction has also successfully been applied to lifting monocular videos to 4D~\cite{ren2024l4gm}. \\
Yet, none of the existing works in the domain have studied the use-case of inferring \textit{animatable} 3D representations from sparse input images, which is the focus of our work. 

\subsection{Monocular 3D Head Avatar Reconstruction}
A popular task is to reconstruct an animatable 3D head avatar from a monocular input video. Several methods have been proposed based on meshes~\cite{grassal2022nha}, NeRFs~\cite{gafni2021nerface, zielonka2023insta, xu2023avatarmav}, SDFs~\cite{zheng2022imavatar}, points~\cite{zheng2023pointavatar}, and Gaussians~\cite{xiang2024flashavatar, chen2024monogaussianavatar}. However, due to their optimization-based nature, these approaches often exhibit poor viewpoint extrapolation capabilities due to overfitting on the input viewpoint. Very recently, ~\citet{tang2024gaf} proposed to improve the geometric fidelity of reconstructed monocular avatars by incorporating priors from a personalized image diffusion model, still requiring a full monocular input video. 
In contrast, our method requires only 4 input images to reconstruct a complete and animatable 3D head with extensive viewpoint coverage.

\subsection{3D-aware Portrait Animation}
A different line of work focuses on animating portraits in a 3D-aware manner, based on 3D volumes~\cite{drobyshev2022megaportraits},  neural rendering~\cite{tewari2020stylerig}, meshes~\cite{khakhulin2022rome}, TriPlanes~\cite{deng2024portrait4d, deng2024portrait4dv2, chu2024gpavatar, tran2024voodoo3d, tran2024voodooxp}, or 3D Gaussians~\cite{chu2024gagavatar}. %
Typically, the animation control is injected via a 3D morphable model (3DMM)~\cite{blanz19993dmm, li2017flame}.
Concurrent to our work, CAP4D~\cite{taubner2024cap4d} uses personalized multi-view image diffusion and LAM~\cite{he2025lam} uses cross attention to build a 3D head avatar from a single image.
While the aforementioned methods can animate a single portrait, they mostly focus on image synthesis from a frontal camera, often trading 3D consistency for better image quality by using 2D screen-space neural renderers. In contrast, our work aims to produce a truthful and complete 3D avatar representation from the input images that can be viewed from any angle.  

\subsection{Photo-realistic 3D Face Models}
The increasing availability of large-scale multi-view face datasets~\cite{kirschstein2023nersemble, ava256, pan2024renderme360, yang2020facescape, buehler2024cafca} has enabled building photo-realistic 3D face models that learn a detailed prior over both geometry and appearance of human faces. HeadNeRF~\cite{hong2022headnerf} conditions a Neural Radiance Field (NeRF)~\cite{mildenhall2021nerf} on identity, expression, albedo, and illumination codes. VRMM~\cite{yang2024vrmm} builds a high-quality and relightable 3D face model using volumetric primitives~\cite{lombardi2021mvp}. One2Avatar~\cite{yu2024one2avatar} extends a 3DMM by anchoring a radiance field to its surface. More recently, GPHM~\cite{xu2025gphm} and HeadGAP~\cite{zheng2024headgap} have adopted 3D Gaussians to build a photo-realistic 3D face model. Another line of work utilizes 3D Generative Adversarial Networks~\cite{chan2021pigan, chan2022eg3d, gu2021stylenerf} to learn the distribution of animatable head avatars~\cite{sun2023next3d, tang20233dfaceshop, lee2022expgan, bergman2022gnarf} from frontal 2D image datasets. \\
To obtain the final 3D avatar, typically a costly fitting process is necessary, impeding casual use-cases on consumer-grade devices.
While this limitation may be circumvented by learning a generalized encoder that maps images into the 3D face model's latent space~\cite{zhao2024invertavatar, sun2022ide3d, liu20223dfmgan, yuan2023goae, bhattarai2024triplanenet}, another fundamental limitation remains. Even with more multi-view face datasets being published, the number of available training subjects rarely exceeds the thousands, making it hard to truly learn the full distribution of human facial appearance. 
3DGAN-based approaches mitigate this by training on single image datasets, but typically struggle with view extra\-polation. 
In this paper, we pursue a different approach that avoids generalizing over the identity axis, by conditioning on some images of a person, and only generalizes over the expression axis, for which plenty of data is available. 

A similar motivation has inspired recent work on codec avatars where a generalized network infers an animatable 3D representation given a registered mesh of a person~\cite{cao2022authentic, li2024uravatar}.
The resulting avatars exhibit excellent quality at the cost of several minutes of video capture per subject and expensive test-time optimization.
For example, URAvatar~\cite{li2024uravatar} finetunes their network on the given video recording for 3 hours on 8 A100 GPUs, making inference on consumer-grade devices impossible. In contrast, our approach directly regresses the final 3D head avatar from just four input images without the need for expensive test-time fine-tuning.

\begin{figure*}[htb]
    \centering
    \includegraphics[width=\textwidth]{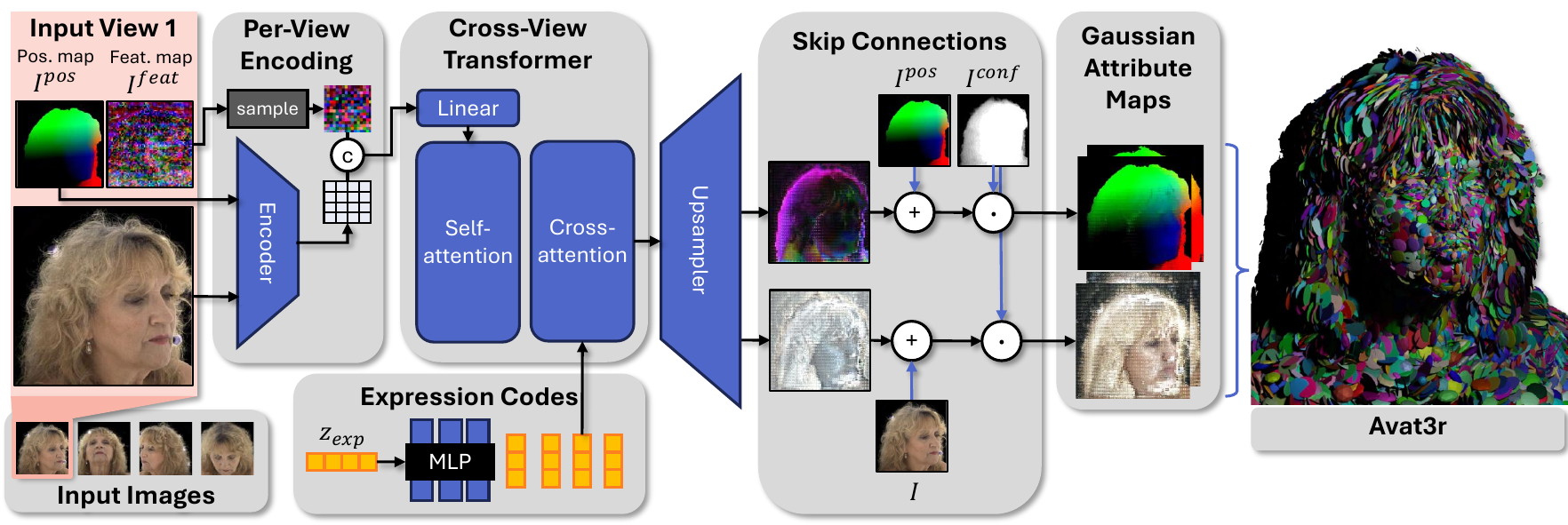}
    \caption{\textbf{Method Overview:} {\OURS} reconstructs a high-quality 3D representation from just a few input images by predicting 3D Gaussian attributes for each input pixel. We first obtain position maps $I^{pos}$ from DUSt3R and feature maps $I^{feat}$ from Sapiens for each view. These are then patchified
    to serve as image tokens for the Vision Transformer back-end. Dense Self-attention performs matching within tokens of the same image and across views to infer 3D structure. Dynamics are modelled via cross-attention layers that attend to a sequenced expression code. The resulting intermediate feature maps are decoded into Gaussian position, scale, rotation, color, and opactiy maps, and then upsampled to the original input image resolution. Finally, we add skip connections to the predicted position and color maps. Gaussians that belong to pixels with a confidence larger than a pre-defined threshold are collected and can be rendered from arbitrary viewpoints. }
    \label{fig:03_method_figure}
\end{figure*}

\section{Method}

The goal of {\OURS} is to predict a high-quality 3D Gaussian representation of a human head from just a few posed input images while simultaneously animating the 3D head to match a desired facial expression:

\begin{align}
    \mathcal{G} = \textsc{Avat3r}(I, \pi, z_{exp})
\end{align}

where $I = \{I_1, ..., I_V\}$ are the $V$ input images, $\pi$ their corresponding camera parameters, and $z_{exp}$ a code describing the desired facial expression, such as an expression code of FLAME~\cite{li2017flame}. In our experiments, $V = 4$. Note that the images $I$ do not necessarily have to stem from multi-view camera systems or synthetic multi-view renderings of objects as it is common for large reconstruction models. Instead, the input images may show the scene at different instances in time, making the model amenable for training and inference on frames from monocular videos. \\
Under the hood, {\OURS} employs a Vision Transformer backbone~\cite{dosovitskiy2020vit} to predict 3D Gaussian primitives for each pixel in the input images. \cref{fig:03_method_figure} depicts the model architecture. In the following, we introduce the different components of our pipeline. 

\subsection{Pre-trained Foundation Models}
We simplify the task by employing two foundation models: DUSt3R~\cite{wang2024dust3r}, and Sapiens~\cite{khirodkar2025sapiens}. DUSt3R predicts dense position maps for a set of input images and globally aligns them with the known camera parameters $\pi$. These position maps will be beneficial for placing the 3D Gaussians. The Sapiens backbone predicts a rich low-resolution feature map for each image. The semantic nature of the extracted features will simplify the view matching task that has to be performed by the subsequent Transformer. 
\begin{align}
    I^{pos}, I^{conf} &= \textsc{Dust3r}(I, \pi) \\
    I^{feat} &= \textsc{Sapiens}(I)
\end{align}
where $I^{pos}, I^{conf}$ are position and confidence maps predicted by DUSt3R, and $I^{feat}$ are the Sapiens feature maps.

\subsection{Animatable Large 3D Reconstruction Model}
We adopt a Vision Transformer architecture following GRM~\cite{xu2024grm} by first patchifying the input maps $I, I^{pos}, I^{pluck} \in \mathbb{R}^{V \times H \times W \times C}$ with a convolutional layer: 
\begin{align}
    h &= \textsc{Conv}\left([I, I^{pos}, I^{pluck}]\right)
\end{align}
where $I^{pluck}$ are the per-ray plucker coordinates stored for each pixel and $h \in \mathbb{R}^{V \times H_p \times W_p \times D}$ will be the low-resolution intermediate feature maps for each input view.\\
Next, we incorporate the pre-computed per-image feature maps $I^{feat} \in \mathbb{R}^{V \times H_f \times W_f \times C_f}$:
\begin{align}
    h^{feat} &= \textsc{GridSample}\left(I^{feat}\right) \\
    h &\leftarrow \textsc{Linear}([h, h^{feat}])
\end{align}
where the grid sampling step is necessary to transform the Sapiens features maps $I^{feat}$ to the same $H_p \times W_p$ resolution as the intermediate feature maps $h$. The subsequent linear layer then squashes the concatenated feature maps back to dimension $D$ to avoid excessive compute increase in the following transformer layers. \\
The core of the Vision Transformer consists of several self-attention layers that perform cross-view matching between all $V\times H_p \times W_p$ tokens to infer 3D information from the given input images:
\begin{align}
    h \leftarrow \textsc{SelfAtt}(h, h)
\end{align} 
To make {\OURS} animatable, we employ several cross-attention layers that allow each image token to attend to a sequence of expression tokens following~\cite{kirschstein2024diffusionavatars, ye2023ipadapter}:
\begin{align}
    f_{exp} &= \textsc{MLP}(z_{exp}) \\
    h &\leftarrow \textsc{CrossAtt}(h, f_{exp})
\end{align}
where the \textsc{MLP} projects a single descriptive expression code $z_{exp} \in \mathbb{R}^{C_{exp}}$ to the expression code sequence $f_{exp} \in \mathbb{R}^{S \times D}$. In our case, $S = 4$. Surprisingly, by simply enriching each image token in $h$ with expression-dependent information, {\OURS} is already capable of modeling complex facial animations in 3D.  \\
We upsample the image tokens again to the original input resolution using the transformer-based upsampler from GRM~\cite{xu2024grm}:
\begin{align}
    M = \textsc{Upsample}(h)
\end{align}
where $M \in \mathbb{R}^{V \times H \times W \times C_{G}}$ denote the Gaussian attribute maps consisting of per-pixel Gaussian positions, scales, rotations, colors, and opacities. After upsampling, we add two skip connections for positions and colors:
\begin{align}
    M^{pos} &\leftarrow M^{pos} + I^{pos} \\
    M^{rgb} &\leftarrow M^{rgb} + I
\end{align}
which incorporate the inductive bias that a pixel should primarily spawn a 3D Gaussian that models the pixel's corresponding 3D geometry. \\
Finally, we make use of DUSt3R's per-pixel confidence maps $I^{conf}$ to decide which pixel in the generated Gaussian attribute maps $M$ should actually spawn a 3D Gaussian:
\begin{align}
    \mathcal{G} = \{M[x, y] : I^{conf}[x, y] > \tau\}
\end{align}
where $\tau = 0.5$ is a threshold describing the minimal confidence needed to create a 3D Gaussian. This serves two purposes: First, it avoids artifacts that arise from incorrect 3D position initializations from DUSt3R. Second, it naturally adjusts the number of Gaussians to the person at hand. In that sense, the confidence mask acts much like a foreground segmentation: If the person had large voluminous hair, then more pixels will belong to the foreground and thus more Gaussians will be kept. \\
The final set of 3D Gaussians $\mathcal{G}$ can then be rendered from any desired novel viewpoint $\pi^{nv}$ using the tile-based differentiable rasterizer $\mathcal{R}$~\cite{kerbl20233dgs}:
\begin{align}
    I^{nv} = \mathcal{R}(\mathcal{G}, \pi^{nv})
\end{align}

\subsection{Losses}

{\OURS} can be trained with only photometric losses on novel viewpoints following the original 3D Gaussian Splatting work~\cite{kerbl20233dgs}:
\begin{align}
    \mathcal{L}_{l1} &= \Vert I^{nv} - I^{gt} \Vert_1 \\
    \mathcal{L}_{ssim} &= \textsc{SSIM}(I^{nv}, I^{gt})
\end{align}

Furthermore, we incorporate perceptual losses~\cite{zhang2018lpips} to encourage the emergence of more high-frequency details:
\begin{align}
    \mathcal{L}_{lpips} &= \textsc{LPIPS}(I^{nv}, I^{gt})
\end{align}

The final loss term is comprised as follows:
\begin{align}
    \mathcal{L} = \lambda_{l1} \mathcal{L}_{l1} + \lambda_{ssim} \mathcal{L}_{ssim} + \lambda_{lpips} \mathcal{L}_{lpips}
\end{align}
with $\lambda_{l1} = 0.8, \lambda_{ssim} = 0.2$, and $\lambda_{lpips} = 0.01$.

\section{Experimental Results}

\subsection{Training}

We train {\OURS} on multi-view video captures from the Ava-256 dataset~\cite{ava256}. In particular, we use the 4TB version of the dataset which contains 7.5fps videos from 80 cameras (roughly 5000 frames each) at a resolution of $1024 \times 667$. {\OURS} is trained on 244 persons from the dataset with the remaining 12 being held out. We use the provided tracked mesh to crop each image to a head-centric $512\times512$ portrait which will be fed as input to our model. For each training sample, we randomly select 4 reasonably diverse viewpoints and 4 timesteps with diverse expressions as input images $I$. On the other hand, the supervision images $I^{gt}$ are left at the highest resolution and are randomly sampled over all available cameras and timesteps. For $z_{exp}$, we use the expression codes that are provided with the Ava256 dataset.\\
{\OURS} is trained with Adam~\cite{kingma2014adam} using a learning rate of $5\text{e-}5$. We only introduce $\mathcal{L}_{lpips}$ after 3M optimization steps to avoid focusing on high-frequency details too early. In total, the model is trained for 3.5M steps with a batch size of 1 per GPU on 8 A100 GPUs which takes roughly 4 days. We further find that training convergence is significantly improved when supervising multiple viewpoints for each batch. I.e., a batch consists of 12 images: 4 with random expression that are used as input, and 8 viewpoints from the target expression that are used as supervision.

\subsection{Experiment Setup}

\paragraph{Task.} 
We evaluate the ability to create a 3D head avatar from four images of an unseen person. 

\paragraph{Metrics.} 
We employ three paired-image metrics to measure the quality of individual rendered images: Peak Signal-to-Noise Ratio (PSNR), Structural Similarity Index (SSIM)~\cite{wang2004ssim}, and Learned Perceptual Image Patch Similarity (LPIPS)~\cite{zhang2018lpips}. 
Furthermore, we make use of two face-specific metrics: Average Keypoint Distance (AKD) measured in pixels with keypoints estimated from PIPNet~\cite{jin2021pipnet}, and cosine similarity (CSIM) of identity embeddings based on ArcFace~\cite{deng2019arcface}.

\paragraph{Baselines.}
We compare our method with recent state-of-the-art systems for 3D head avatar creation from few input images. 

\textit{HeadNeRF~\cite{hong2022headnerf}.} A NeRF-based Autodecoder with disentangled identity and expression spaces that can be fitted to an arbitrary number of images.

\textit{InvertAvatar~\cite{zhao2024invertavatar}.} A feed-forward encoder that predicts the underlying representation of Next3D~\cite{sun2023next3d} from 4 input images which then can be animated.

\textit{GPAvatar~\cite{chu2024gpavatar}.} A NeRF-based method that predicts canonical TriPlanes from multiple input images. Animations are then modeled by querying 3D points in an expression-dependent feature field inferred from FLAME.

\begin{figure*}
    \centering
    \begin{tabularx}{0.05\linewidth}{l}
        \rotatebox[origin=r]{90}{\parbox[c]{5.3cm}{\centering $\overbrace{\hspace{5.3cm}}^{\text{\normalsize Ava256 dataset}}$}} \\
        \rotatebox[origin=r]{90}{\parbox[c]{5.3cm}{\centering $\overbrace{\hspace{5.3cm}}^{\text{\normalsize NeRSemble dataset}}$}}
    \end{tabularx}%
    \includegraphics[width=0.95\linewidth, valign=c]{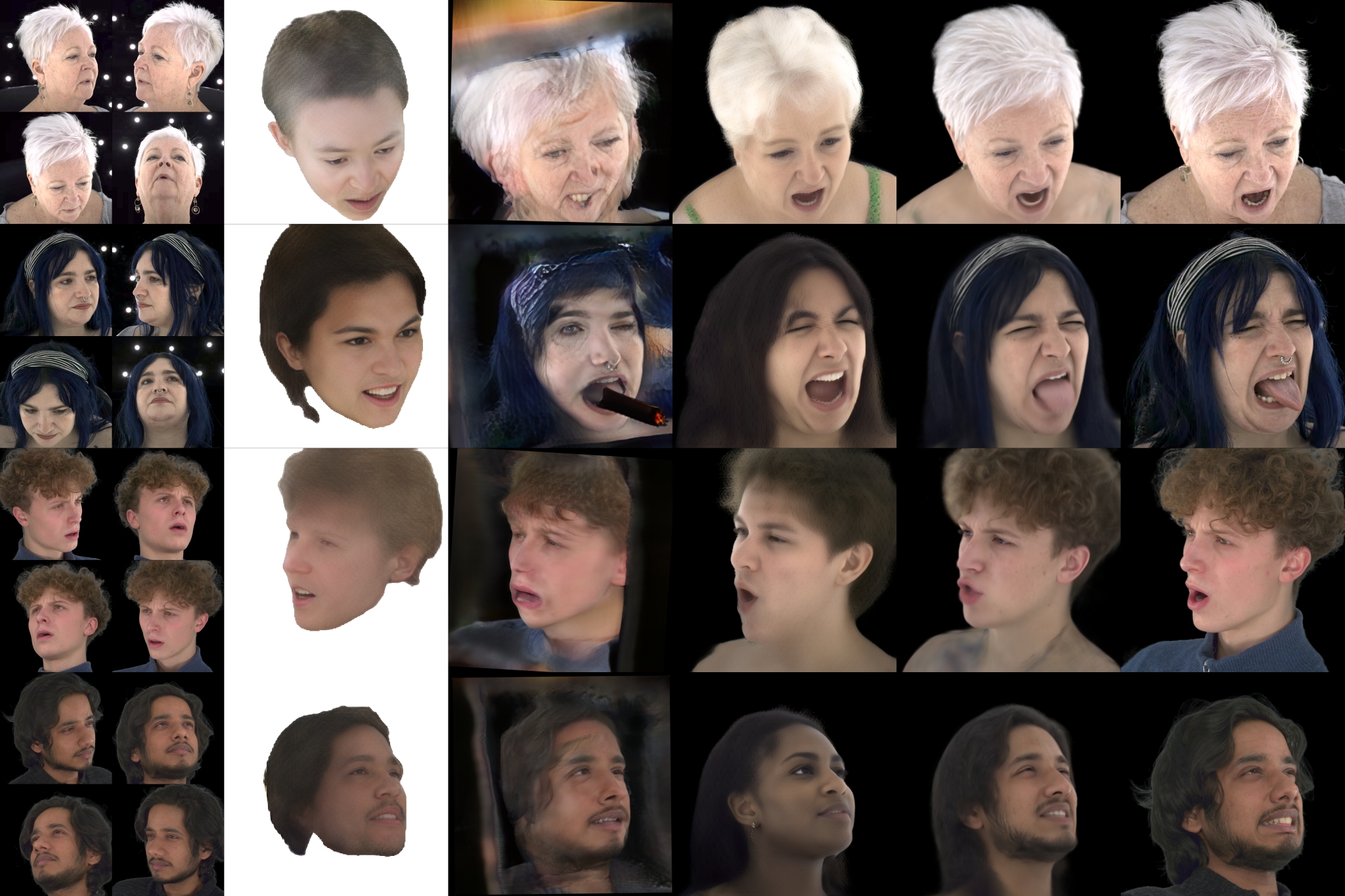}
    \begin{tabularx}{\linewidth}{P{0.025\linewidth}YYYYYY}
        & Input images & HeadNeRF~\cite{hong2022headnerf} & InvertAvatar~\cite{zhao2024invertavatar} & GPAvatar~\cite{chu2024gpavatar} & Ours & GT
    \end{tabularx}
    \caption{\textbf{Comparison for few-shot 3D Head Avatar creation.} We evaluate the ability of {\OURS} to create 3D head avatars from four input images in a self-reenactment setting. 
    Note that the NeRSemble dataset has not been used during training of~{\OURS} and therefore both source and driver person are out-of-domain in this case. Nevertheless, our method produces high quality avatars.}
    \label{fig:4_few_shot_comparison}
\end{figure*}

\subsection{Few-shot 3D Head Avatar Creation}

We compare the ability to create a realistic 3D head avatar from just four input images on both the Ava256 dataset~\cite{ava256} and the NeRSemble dataset~\cite{kirschstein2023nersemble}. To this end, we train GPAvatar on the Ava256 dataset in the same setting as our model and use the high-quality tracked meshes provided with the dataset to drive GPAvatar. For HeadNeRF and InvertAvatar, we use the published code bases for reconstruction, animation and rendering. We evaluate all methods at $512 \times 512$ pixels resolution on the same hold-out persons. For evaluation on Ava256, we sample 1000 different target expressions from 12 persons from different viewpoints. On NeRSemble, we sample 100 target images from 3 male and 3 female subjects. \\
The quantitative results are listed in~\cref{tab:4_results}. It can be seen that our method clearly outperforms the baselines.
We observe large differences in metrics measuring alignment (PSNR, SSIM), identity similarity (CSIM), and facial keypoint similarity (AKD). This indicates that {\OURS} produces 3D head avatars that better resemble the person in the input images and produces more realistic animations. Note that HeadNeRF cannot render any other background than white. We therefore compute metrics against ground truth images with white background for that method. We further show qualitative comparisons in~\cref{fig:4_few_shot_comparison}. Our method produces more expressive avatars with higher rendering quality than the baselines and even performs well on a person with colored hair who wears a headband. Also, note that the NeRSemble dataset has not been used during training and the images in this dataset exhibit entirely different characteristics in terms of lighting, camera parameters, and viewpoints, compared to the training images from the Ava256 dataset. Nevertheless, {\OURS} generates realistic 3D head avatars in this setting as well. \\
For video renderings and single-image results, see our supplemental material.

\begin{table}[htb]
    \setlength{\tabcolsep}{3pt}
    \centering
    \begin{tabular}{llrrrrr}
        \toprule
            && \footnotesize{PSNR}$\uparrow$ & \footnotesize{SSIM}$\uparrow$ & \footnotesize{LPIPS}$\downarrow$ & \footnotesize{AKD}$\downarrow$ & \footnotesize{CSIM}$\uparrow$ \\

        \midrule
            \multirow{4}{*}{\rotatebox[origin=c]{90}{\parbox[c]{1.5cm}{\centering \small Ava256}}}
                & HeadNeRF %
                    & 9.1 & 0.64 & 0.52 & 6.9 & 0.11 \\
                & InvertAvatar %
                    & 14.2 & 0.36 & 0.55 & 15.8 & 0.29 \\
                & GPAvatar %
                    & 19.4 & 0.69 & 0.34 & 5.3 & 0.31 \\
                & Ours  %
                    & \textbf{20.7} & \textbf{0.71} & \textbf{0.33} & \textbf{4.8} & \textbf{0.59}
                    \\
        \midrule
            \multirow{4}{*}{\rotatebox[origin=c]{90}{\parbox[c]{1.5cm}{\centering \small NeRSemble}}}
                & HeadNeRF
                    & 9.8           & 0.69          & 0.47          & 4.9          & 0.22 \\
                & InvertAvatar
                    & 17.2          & 0.44          & 0.49          & 5.6          & 0.44 \\
                & GPAvatar\textsuperscript{\textdagger} %
                    & 17.6          & 0.67          & 0.40          & 5.7          & 0.07 \\ %
                & Ours 
                    & \textbf{20.5} & \textbf{0.75} & \textbf{0.33} & \textbf{3.7} & \textbf{0.50} \\
        \bottomrule
    \end{tabular}
    \caption{\textbf{3D Head Avatar Creation Comparison.} Our method outperforms the baselines in all metrics.
    }
    \label{tab:4_results}
\end{table}

\begin{figure*}
    \centering
    \includegraphics[width=\linewidth]{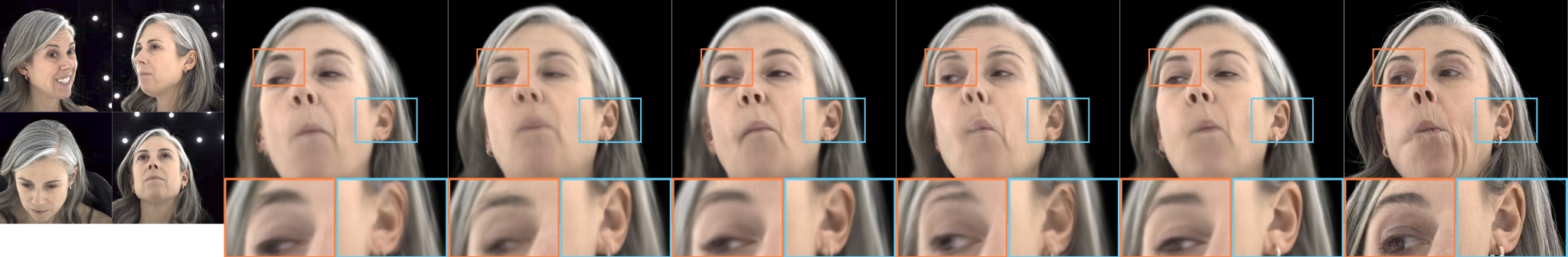}
    \begin{tabularx}{\linewidth}{YYYYYYYY}
        Input Images & (a) Dynamic GRM & (b) w/o DUSt3R & (c) w/o Sapiens & (d) w/o rand. timesteps & (e) Ours full & GT
    \end{tabularx}
    \caption{\textbf{Ablation study.} Adding Sapiens features (b) noticeably improves sharpness but still exhibits misalignments. Conversely, using position maps from Dust3r improves alignment but is overall less sharp (c). Combining both but training on multi-view images from the same timestep produces sharp results but is less robust to inconsistencies in the input (d). Our final model exhibits the best trade-off between alignment, sharpness, and robustness (e).  }
    \label{fig:4_ablation}
\end{figure*}

\newcommand{\checkbox}{\scalebox{1.5}{$\boxtimes$}}
\newcommand{\emptybox}{\scalebox{1.5}{$\square$}}

\begin{table}[tb]
    \setlength{\tabcolsep}{1pt}
    \centering
    \begin{tabular}{lcccrrrrr}
        \toprule
             & $\mathcal{S}$ & $\mathcal{D}$ & $\mathcal{T}$ 
             & {\small PSNR}$\uparrow$ 
             && {\small LPIPS}$\downarrow$
             && {\small AKD}$\downarrow$
             \\
        \midrule
            Dynamic GRM & \emptybox & \emptybox & \checkbox
                & 20.8 && 0.439 && 8.60 \\ %
            w/o Sapiens & \emptybox & \checkbox & \checkbox
                & 20.9 && 0.434 && 8.59 \\ %
            w/o Dust3r & \checkbox & \emptybox & \checkbox
                & 21.1 && 0.429 && 9.60 \\ %
            w/o rand. timesteps & \checkbox & \checkbox & \emptybox
                & 21.3 && \textbf{0.409} && 8.86 \\ %
            Avat3r & \checkbox & \checkbox & \checkbox
                & \textbf{21.6} && 0.410 && \textbf{8.08}\\ %
        \bottomrule
    \end{tabular}
    \caption{\textbf{Ablation.} We analyze the effect of Sapiens features~($\mathcal{S}$), DUSt3R position maps~($\mathcal{D}$), and training on randomized input timesteps~($\mathcal{T}$). All ablation models are trained without LPIPS loss. Metrics are computed on $667 \times 667$ renderings.}
    \label{tab:4_ablations}
\end{table}

\subsection{Ablations}

In the following, we study the efficacy of our design choices for Avat3r. In particular, we look at the usefulness of DUSt3R's position maps, the Sapiens feature maps, and using input images with different expressions during training. More ablations and a runtime analysis can be found in the supplemental material. Our findings are summarized in~\cref{tab:4_ablations} and~\cref{fig:4_ablation}.

\paragraph{Effect of DUSt3R position maps.} Removing DUSt3R from our pipeline mostly impairs geometric fidelity. In~\cref{fig:4_ablation}, it can be seen that {\OURS} without DUSt3R struggles with aligning the Gaussian predictions from the four input images.

\paragraph{Effect of Sapiens feature maps.} Sapiens provides rich semantic information for the input images that mostly help in producing sharper predictions, e.g., in the hair area.

\paragraph{Effect of training with inconsistent input images.} A model that is trained on only multi-view consistent input images produces slightly sharper images. However, any inconsistencies in the input images, which can easily happen in practice, carry over to the generated 3D representation causing unwanted artifacts.

\subsection{Additional Results}  %

\begin{figure}[htb]
    \setlength{\tabcolsep}{0pt}
    \centering
    \includegraphics[width=\linewidth]{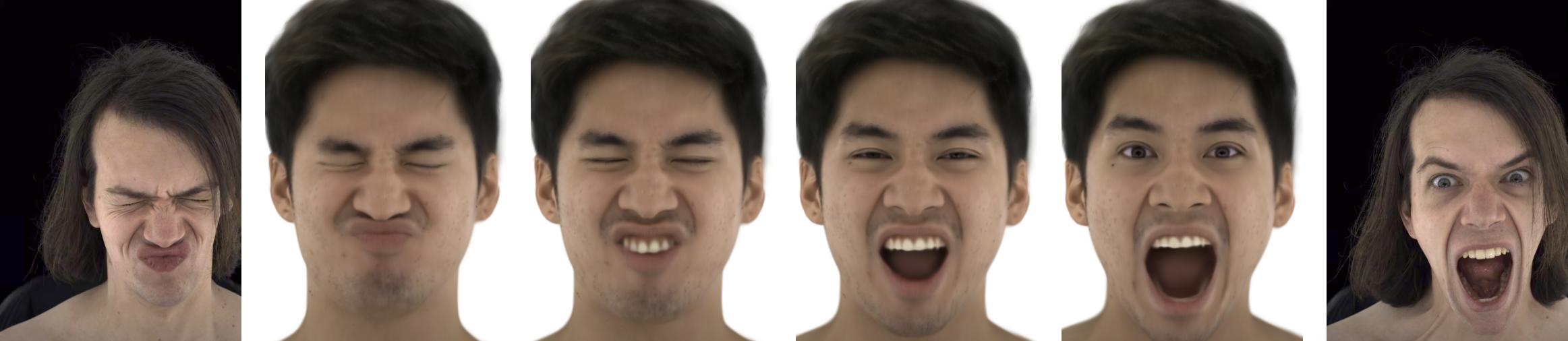}
    \begin{tabularx}{\linewidth}{P{0.16\linewidth}P{0.68\linewidth}P{0.16\linewidth}}
        \multicolumn{3}{c}{\vspace{-0.5cm}}\\
        \footnotesize{Driver 1} & $\overrightarrow{\hspace{0.25\linewidth}\text{\footnotesize{Expression Interpolation}}\hspace{0.25\linewidth}}$ & \footnotesize{Driver 2}
    \end{tabularx}
    \vspace{-0.5cm}
    \caption{\textbf{Expression Generalization.} 
    }
    \vspace{-0.5cm}

    \label{fig:expression_interpolation}
\end{figure}

\paragraph{Extreme expressions.}
We showcase the generalization capability of our method by interpolating between two extreme expressions 
in~\cref{fig:expression_interpolation}. Both source and driver person are unseen.

\begin{figure*}[htb]
    \setlength{\tabcolsep}{0pt}
    \centering
    \includegraphics[width=\linewidth]{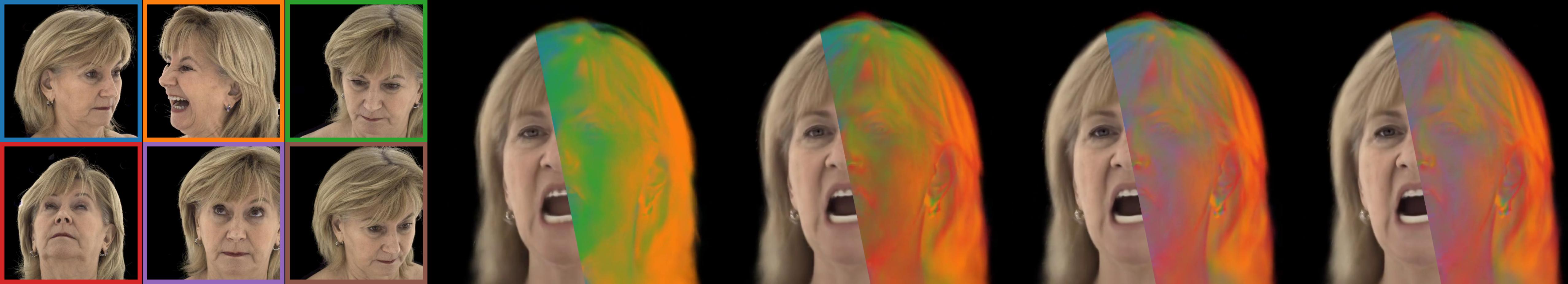}
    \begin{tabularx}{\linewidth}{P{0.27\linewidth}P{0.19\linewidth}P{0.18\linewidth}P{0.19\linewidth}P{0.18\linewidth}}
        \multicolumn{5}{c}{\vspace{-1cm}}\\
        \multicolumn{5}{r}{$\overrightarrow{\vspace{-0.25cm}\hspace{0.72\linewidth}}\hspace{3pt}$}\\
        \footnotesize{input images} & \footnotesize{3} & \footnotesize{4 (train)} & \footnotesize{5} & \footnotesize{6} \\
    \end{tabularx}
    \vspace{-0.6cm}
    \caption{\textbf{Input Flexibility.} Our model accepts different numbers of views. Colors indicate from which view a Gaussian originates.}
    \vspace{-0.3cm}
    \label{fig:ablation_input_views}

\end{figure*}

\paragraph{Adapting to arbitrary input views.}
Despite only being trained with 4 input views, our model can be directly used when more or fewer views are available, see~\cref{fig:ablation_input_views}. 

\begin{figure}
    \centering
    \includegraphics[width=\linewidth]{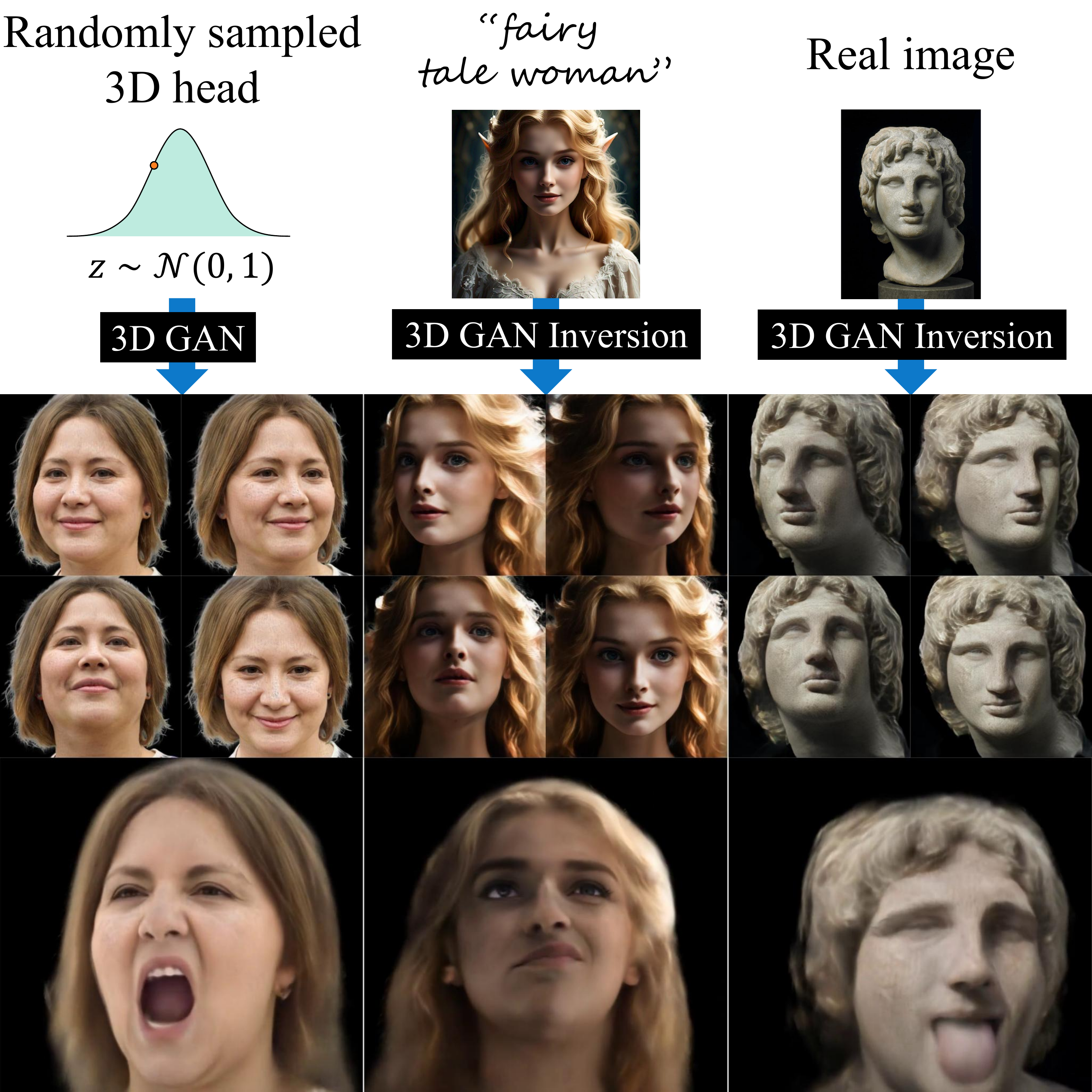}
    \begin{tabularx}{\linewidth}{P{0.3\linewidth}P{0.28\linewidth}Y}
        (a) Animate sampled 3D head & (b) Animate a generated image & (c) Animate photo of a bust
    \end{tabularx}
    \caption{\textbf{Application scenarios of {\OURS}.} We demonstrate various use-cases of our model such as animating a randomly sampled 3D head, animating an image generated by a text-to-image diffusion model, and animating a picture of a greek bust.}
    \vspace{-0.7em}
    \label{fig:4_applications}
\end{figure}

\subsection{Applications}

To make Avat3r amenable for inference on only a single input image, we make use of a pre-trained 3D GAN~\cite{yuan2023goae} to first lift the single image to 3D and then render four views of the head. In~\cref{fig:4_applications}, we show three use-cases of Avat3r: (i) Generating a novel 3D head with GGHead~\cite{kirschstein2024gghead} and then animating it with {\OURS}, (ii) Generating a portrait image with a text-to-image diffusion model~\cite{dai2023emu}, lifting it with a 3D GAN~\cite{yuan2023goae}, and animating it with {\OURS}, and (iii) Lifting an existing photograph with a 3D GAN~\cite{yuan2023goae} and animating it with {\OURS}. The results demonstrate that, despite being designed for a few-shot scenario and only being trained on real studio data, Avat3r can be used in a variety of casual situations. For example, our findings for case (i) suggest that Avat3r may be used to animate images from any 3D GAN that generates 3D heads, effectively making any purely static 3D generation model animatable.

\subsection{Limitations}

Despite {\OURS}'s great ability to create 3D head avatars from a few or even just a single input image, several limitations remain. For example, the current pipeline for inferring avatars from a single image requires a 3D GAN for 3D lifting which accumulates errors in the process in two ways: First, the 3D GAN cannot perfectly explain the person, inevitably loosing some details, and second, existing inversion frameworks employ NeRF-based 3D GANs that rely on screen-space super-resolution after rendering which introduces view-inconsistencies to the final images. Both aspects hamper the quality of the final avatar. However, these issues will decrease as newer and better 3D GANs become available that do not require screen-space rendering anymore~\cite{kirschstein2024gghead, Trevithick2023wysiwyg}. 
Another limitation of our framework is the requirement of camera poses during inference. While reasonable estimates for face images can be obtained by head tracking algorithms~\cite{MICA:ECCV2022}, incorrect camera estimates still remain a potential source of error for real use-cases. Possible remedies may include teaching the network to deal with incorrect camera estimates already during training, or dropping the requirement for camera poses alltogether~\cite{wang2023pflrm}.
Finally, our current pipeline does not provide control over the lighting and instead bakes in light effects from the input images. This limits applications that put avatars into novel environments since the generated head will look out of place. Adressing this issue by explicitly disentangling the lighting properties is an exciting avenue for future research.

\section{Conclusion}
We have presented {\OURS}, a system for directly regressing an animatable 3D head avatar from just four input images. Our pipeline combines recent advancements in large reconstruction models with the powerful foundation models DUSt3R and Sapiens to obtain high quality results, and models dynamics with simple cross attention layers. Moreover, we demonstrate that {\OURS} can also be employed in a single image scenario by using a pre-trained 3D GAN. Our experiments show that {\OURS} learns a powerful prior over 3D facial movement that even generalizes to out-of-distribution examples like pictures of antique busts or AI-generated images. We believe that our findings open up interesting avenues for future research. For example, we have shown that LRM-like architectures have the ability to infer 3D representations even from inconsistent input images such as frames of a monocular video. This opens up the possibility to train models on large-scale monocular video datasets with a much larger variety of identities. Finally, the underlying architecture could also be used as a 3D lifting and denoising module in a diffusion framework~\cite{anciukevivcius2023renderdiffusion, chan2023genvs}, enabling training of diffusion models on monocular video data for unconditional 3D head avatar generation.

\subsection*{Acknowledgements}
This work was supported by the ERC Starting Grant Scan2CAD (804724) and the German Research Foundation (DFG) Research Unit ``Learning and Simulation in Visual Computing''.
We would also like to thank Angela Dai for the video voice-over.

{
    \small
    \bibliographystyle{ieeenat_fullname}
    \bibliography{main}

\begin{thebibliography}{86}
\providecommand{\natexlab}[1]{#1}
\providecommand{\url}[1]{\texttt{#1}}
\expandafter\ifx\csname urlstyle\endcsname\relax
  \providecommand{\doi}[1]{doi: #1}\else
  \providecommand{\doi}{doi: \begingroup \urlstyle{rm}\Url}\fi

\bibitem[Anciukevi{\v{c}}ius et~al.(2023)Anciukevi{\v{c}}ius, Xu, Fisher,
  Henderson, Bilen, Mitra, and Guerrero]{anciukevivcius2023renderdiffusion}
Titas Anciukevi{\v{c}}ius, Zexiang Xu, Matthew Fisher, Paul Henderson, Hakan
  Bilen, Niloy~J Mitra, and Paul Guerrero.
\newblock Renderdiffusion: Image diffusion for 3d reconstruction, inpainting
  and generation.
\newblock In \emph{Proceedings of the IEEE/CVF conference on computer vision
  and pattern recognition}, pages 12608--12618, 2023.

\bibitem[Bergman et~al.(2022)Bergman, Kellnhofer, Yifan, Chan, Lindell, and
  Wetzstein]{bergman2022gnarf}
Alexander Bergman, Petr Kellnhofer, Wang Yifan, Eric Chan, David Lindell, and
  Gordon Wetzstein.
\newblock Generative neural articulated radiance fields.
\newblock \emph{Advances in Neural Information Processing Systems},
  35:\penalty0 19900--19916, 2022.

\bibitem[Bhattarai et~al.(2024)Bhattarai, Nie{\ss}ner, and
  Sevastopolsky]{bhattarai2024triplanenet}
Ananta~R Bhattarai, Matthias Nie{\ss}ner, and Artem Sevastopolsky.
\newblock Triplanenet: An encoder for eg3d inversion.
\newblock In \emph{Proceedings of the IEEE/CVF Winter Conference on
  Applications of Computer Vision}, pages 3055--3065, 2024.

\bibitem[Blanz and Vetter(1999)]{blanz19993dmm}
V Blanz and T Vetter.
\newblock A morphable model for the synthesis of 3d faces.
\newblock In \emph{26th Annual Conference on Computer Graphics and Interactive
  Techniques (SIGGRAPH 1999)}, pages 187--194. ACM Press, 1999.

\bibitem[Buehler et~al.(2024)Buehler, Li, Wood, Helminger, Chen, Shah, Wang,
  Garbin, Orts-Escolano, Hilliges, et~al.]{buehler2024cafca}
Marcel~C Buehler, Gengyan Li, Erroll Wood, Leonhard Helminger, Xu Chen, Tanmay
  Shah, Daoye Wang, Stephan Garbin, Sergio Orts-Escolano, Otmar Hilliges,
  et~al.
\newblock Cafca: High-quality novel view synthesis of expressive faces from
  casual few-shot captures.
\newblock In \emph{SIGGRAPH Asia 2024 Conference Papers}, pages 1--12, 2024.

\bibitem[Cao et~al.(2022)Cao, Simon, Kim, Schwartz, Zollhoefer, Saito,
  Lombardi, Wei, Belko, Yu, et~al.]{cao2022authentic}
Chen Cao, Tomas Simon, Jin~Kyu Kim, Gabriel Schwartz, Michael Zollhoefer,
  Shun-Suke Saito, Stephen Lombardi, Shih-En Wei, Danielle Belko, Shoou-I Yu,
  et~al.
\newblock Authentic volumetric avatars from a phone scan.
\newblock 2022.

\bibitem[Chan et~al.(2021)Chan, Monteiro, Kellnhofer, Wu, and
  Wetzstein]{chan2021pigan}
Eric~R Chan, Marco Monteiro, Petr Kellnhofer, Jiajun Wu, and Gordon Wetzstein.
\newblock pi-gan: Periodic implicit generative adversarial networks for
  3d-aware image synthesis.
\newblock In \emph{Proceedings of the IEEE/CVF conference on computer vision
  and pattern recognition}, pages 5799--5809, 2021.

\bibitem[Chan et~al.(2022)Chan, Lin, Chan, Nagano, Pan, De~Mello, Gallo,
  Guibas, Tremblay, Khamis, et~al.]{chan2022eg3d}
Eric~R Chan, Connor~Z Lin, Matthew~A Chan, Koki Nagano, Boxiao Pan, Shalini
  De~Mello, Orazio Gallo, Leonidas~J Guibas, Jonathan Tremblay, Sameh Khamis,
  et~al.
\newblock Efficient geometry-aware 3d generative adversarial networks.
\newblock In \emph{Proceedings of the IEEE/CVF conference on computer vision
  and pattern recognition}, pages 16123--16133, 2022.

\bibitem[Chan et~al.(2023)Chan, Nagano, Chan, Bergman, Park, Levy, Aittala,
  Mello, Karras, and Wetzstein]{chan2023genvs}
Eric~R. Chan, Koki Nagano, Matthew~A. Chan, Alexander~W. Bergman, Jeong~Joon
  Park, Axel Levy, Miika Aittala, Shalini~De Mello, Tero Karras, and Gordon
  Wetzstein.
\newblock {GeNVS}: Generative novel view synthesis with {3D}-aware diffusion
  models.
\newblock In \emph{arXiv}, 2023.

\bibitem[Charatan et~al.(2024)Charatan, Li, Tagliasacchi, and
  Sitzmann]{charatan2024pixelsplat}
David Charatan, Sizhe~Lester Li, Andrea Tagliasacchi, and Vincent Sitzmann.
\newblock pixelsplat: 3d gaussian splats from image pairs for scalable
  generalizable 3d reconstruction.
\newblock In \emph{Proceedings of the IEEE/CVF Conference on Computer Vision
  and Pattern Recognition}, pages 19457--19467, 2024.

\bibitem[Chen et~al.(2024)Chen, Wang, Li, Xiao, Zhang, Yao, and
  Liu]{chen2024monogaussianavatar}
Yufan Chen, Lizhen Wang, Qijing Li, Hongjiang Xiao, Shengping Zhang, Hongxun
  Yao, and Yebin Liu.
\newblock Monogaussianavatar: Monocular gaussian point-based head avatar.
\newblock In \emph{ACM SIGGRAPH 2024 Conference Papers}, pages 1--9, 2024.

\bibitem[Chen et~al.(2025)Chen, Xu, Zheng, Zhuang, Pollefeys, Geiger, Cham, and
  Cai]{chen2025mvsplat}
Yuedong Chen, Haofei Xu, Chuanxia Zheng, Bohan Zhuang, Marc Pollefeys, Andreas
  Geiger, Tat-Jen Cham, and Jianfei Cai.
\newblock Mvsplat: Efficient 3d gaussian splatting from sparse multi-view
  images.
\newblock In \emph{European Conference on Computer Vision}, pages 370--386.
  Springer, 2025.

\bibitem[Chu and Harada(2024)]{chu2024gagavatar}
Xuangeng Chu and Tatsuya Harada.
\newblock Generalizable and animatable gaussian head avatar.
\newblock \emph{arXiv preprint arXiv:2410.07971}, 2024.

\bibitem[Chu et~al.(2024)Chu, Li, Zeng, Yang, Lin, Liu, and
  Harada]{chu2024gpavatar}
Xuangeng Chu, Yu Li, Ailing Zeng, Tianyu Yang, Lijian Lin, Yunfei Liu, and
  Tatsuya Harada.
\newblock Gpavatar: Generalizable and precise head avatar from image (s).
\newblock \emph{arXiv preprint arXiv:2401.10215}, 2024.

\bibitem[Dai et~al.(2023)Dai, Hou, Ma, Tsai, Wang, Wang, Zhang, Vandenhende,
  Wang, Dubey, et~al.]{dai2023emu}
Xiaoliang Dai, Ji Hou, Chih-Yao Ma, Sam Tsai, Jialiang Wang, Rui Wang, Peizhao
  Zhang, Simon Vandenhende, Xiaofang Wang, Abhimanyu Dubey, et~al.
\newblock Emu: Enhancing image generation models using photogenic needles in a
  haystack.
\newblock \emph{arXiv preprint arXiv:2309.15807}, 2023.

\bibitem[Deng et~al.(2019)Deng, Guo, Xue, and Zafeiriou]{deng2019arcface}
Jiankang Deng, Jia Guo, Niannan Xue, and Stefanos Zafeiriou.
\newblock Arcface: Additive angular margin loss for deep face recognition.
\newblock In \emph{Proceedings of the IEEE/CVF conference on computer vision
  and pattern recognition}, pages 4690--4699, 2019.

\bibitem[Deng et~al.(2024{\natexlab{a}})Deng, Wang, Ren, Chen, and
  Wang]{deng2024portrait4d}
Yu Deng, Duomin Wang, Xiaohang Ren, Xingyu Chen, and Baoyuan Wang.
\newblock Portrait4d: Learning one-shot 4d head avatar synthesis using
  synthetic data.
\newblock In \emph{Proceedings of the IEEE/CVF Conference on Computer Vision
  and Pattern Recognition}, pages 7119--7130, 2024{\natexlab{a}}.

\bibitem[Deng et~al.(2024{\natexlab{b}})Deng, Wang, and
  Wang]{deng2024portrait4dv2}
Yu Deng, Duomin Wang, and Baoyuan Wang.
\newblock Portrait4d-v2: Pseudo multi-view data creates better 4d head
  synthesizer.
\newblock \emph{arXiv preprint arXiv:2403.13570}, 2024{\natexlab{b}}.

\bibitem[Dosovitskiy et~al.(2021)Dosovitskiy, Beyer, Kolesnikov, Weissenborn,
  Zhai, Unterthiner, Dehghani, Minderer, Heigold, Gelly, Uszkoreit, and
  Houlsby]{dosovitskiy2020vit}
Alexey Dosovitskiy, Lucas Beyer, Alexander Kolesnikov, Dirk Weissenborn,
  Xiaohua Zhai, Thomas Unterthiner, Mostafa Dehghani, Matthias Minderer, Georg
  Heigold, Sylvain Gelly, Jakob Uszkoreit, and Neil Houlsby.
\newblock An image is worth 16x16 words: Transformers for image recognition at
  scale.
\newblock \emph{ICLR}, 2021.

\bibitem[Drobyshev et~al.(2022)Drobyshev, Chelishev, Khakhulin, Ivakhnenko,
  Lempitsky, and Zakharov]{drobyshev2022megaportraits}
Nikita Drobyshev, Jenya Chelishev, Taras Khakhulin, Aleksei Ivakhnenko, Victor
  Lempitsky, and Egor Zakharov.
\newblock Megaportraits: One-shot megapixel neural head avatars.
\newblock In \emph{Proceedings of the 30th ACM International Conference on
  Multimedia}, pages 2663--2671, 2022.

\bibitem[Durvasula et~al.(2023)Durvasula, Zhao, Chen, Liang, Sanjaya, and
  Vijaykumar]{durvasula2023distwar}
Sankeerth Durvasula, Adrian Zhao, Fan Chen, Ruofan Liang, Pawan~Kumar Sanjaya,
  and Nandita Vijaykumar.
\newblock Distwar: Fast differentiable rendering on raster-based rendering
  pipelines.
\newblock \emph{arXiv preprint arXiv:2401.05345}, 2023.

\bibitem[Gafni et~al.(2021)Gafni, Thies, Zollhofer, and
  Nie{\ss}ner]{gafni2021nerface}
Guy Gafni, Justus Thies, Michael Zollhofer, and Matthias Nie{\ss}ner.
\newblock Dynamic neural radiance fields for monocular 4d facial avatar
  reconstruction.
\newblock In \emph{Proceedings of the IEEE/CVF Conference on Computer Vision
  and Pattern Recognition}, pages 8649--8658, 2021.

\bibitem[Giebenhain et~al.(2024)Giebenhain, Kirschstein, R{\"u}nz, Agapito, and
  Nie{\ss}ner]{giebenhain2024npga}
Simon Giebenhain, Tobias Kirschstein, Martin R{\"u}nz, Lourdes Agapito, and
  Matthias Nie{\ss}ner.
\newblock Npga: Neural parametric gaussian avatars.
\newblock \emph{arXiv preprint arXiv:2405.19331}, 2024.

\bibitem[Grassal et~al.(2022)Grassal, Prinzler, Leistner, Rother, Nie{\ss}ner,
  and Thies]{grassal2022nha}
Philip-William Grassal, Malte Prinzler, Titus Leistner, Carsten Rother,
  Matthias Nie{\ss}ner, and Justus Thies.
\newblock Neural head avatars from monocular rgb videos.
\newblock In \emph{Proceedings of the IEEE/CVF conference on computer vision
  and pattern recognition}, pages 18653--18664, 2022.

\bibitem[Gu et~al.(2021)Gu, Liu, Wang, and Theobalt]{gu2021stylenerf}
Jiatao Gu, Lingjie Liu, Peng Wang, and Christian Theobalt.
\newblock Stylenerf: A style-based 3d-aware generator for high-resolution image
  synthesis.
\newblock \emph{arXiv preprint arXiv:2110.08985}, 2021.

\bibitem[He et~al.(2025)He, Gu, Ye, Xu, Zhao, Dong, Yuan, Dong, and
  Bo]{he2025lam}
Yisheng He, Xiaodong Gu, Xiaodan Ye, Chao Xu, Zhengyi Zhao, Yuan Dong, Weihao
  Yuan, Zilong Dong, and Liefeng Bo.
\newblock Lam: Large avatar model for one-shot animatable gaussian head.
\newblock \emph{arXiv preprint arXiv:2502.17796}, 2025.

\bibitem[Hong et~al.(2022)Hong, Peng, Xiao, Liu, and Zhang]{hong2022headnerf}
Yang Hong, Bo Peng, Haiyao Xiao, Ligang Liu, and Juyong Zhang.
\newblock Headnerf: A real-time nerf-based parametric head model.
\newblock In \emph{Proceedings of the IEEE/CVF Conference on Computer Vision
  and Pattern Recognition}, pages 20374--20384, 2022.

\bibitem[Hong et~al.(2023)Hong, Zhang, Gu, Bi, Zhou, Liu, Liu, Sunkavalli, Bui,
  and Tan]{hong2023lrm}
Yicong Hong, Kai Zhang, Jiuxiang Gu, Sai Bi, Yang Zhou, Difan Liu, Feng Liu,
  Kalyan Sunkavalli, Trung Bui, and Hao Tan.
\newblock Lrm: Large reconstruction model for single image to 3d.
\newblock \emph{arXiv preprint arXiv:2311.04400}, 2023.

\bibitem[Jin et~al.(2021)Jin, Liao, and Shao]{jin2021pipnet}
Haibo Jin, Shengcai Liao, and Ling Shao.
\newblock Pixel-in-pixel net: Towards efficient facial landmark detection in
  the wild.
\newblock \emph{International Journal of Computer Vision}, 2021.

\bibitem[Kabadayi et~al.(2024)Kabadayi, Zielonka, Bhatnagar, Pons-Moll, and
  Thies]{kabadayi2024ganavatar}
Berna Kabadayi, Wojciech Zielonka, Bharat~Lal Bhatnagar, Gerard Pons-Moll, and
  Justus Thies.
\newblock Gan-avatar: Controllable personalized gan-based human head avatar.
\newblock In \emph{2024 International Conference on 3D Vision (3DV)}, pages
  882--892. IEEE, 2024.

\bibitem[Kerbl et~al.(2023)Kerbl, Kopanas, Leimk{\"u}hler, and
  Drettakis]{kerbl20233dgs}
Bernhard Kerbl, Georgios Kopanas, Thomas Leimk{\"u}hler, and George Drettakis.
\newblock 3d gaussian splatting for real-time radiance field rendering.
\newblock \emph{ACM Trans. Graph.}, 42\penalty0 (4):\penalty0 139--1, 2023.

\bibitem[Khakhulin et~al.(2022)Khakhulin, Sklyarova, Lempitsky, and
  Zakharov]{khakhulin2022rome}
Taras Khakhulin, Vanessa Sklyarova, Victor Lempitsky, and Egor Zakharov.
\newblock Realistic one-shot mesh-based head avatars.
\newblock In \emph{European Conference on Computer Vision}, pages 345--362.
  Springer, 2022.

\bibitem[Khirodkar et~al.(2025)Khirodkar, Bagautdinov, Martinez, Zhaoen, James,
  Selednik, Anderson, and Saito]{khirodkar2025sapiens}
Rawal Khirodkar, Timur Bagautdinov, Julieta Martinez, Su Zhaoen, Austin James,
  Peter Selednik, Stuart Anderson, and Shunsuke Saito.
\newblock Sapiens: Foundation for human vision models.
\newblock In \emph{European Conference on Computer Vision}, pages 206--228.
  Springer, 2025.

\bibitem[Kingma and Ba(2014)]{kingma2014adam}
Diederik~P Kingma and Jimmy Ba.
\newblock Adam: A method for stochastic optimization.
\newblock \emph{arXiv preprint arXiv:1412.6980}, 2014.

\bibitem[Kirschstein et~al.(2023)Kirschstein, Qian, Giebenhain, Walter, and
  Nie{\ss}ner]{kirschstein2023nersemble}
Tobias Kirschstein, Shenhan Qian, Simon Giebenhain, Tim Walter, and Matthias
  Nie{\ss}ner.
\newblock Nersemble: Multi-view radiance field reconstruction of human heads.
\newblock \emph{ACM Transactions on Graphics (TOG)}, 42\penalty0 (4):\penalty0
  1--14, 2023.

\bibitem[Kirschstein et~al.(2024{\natexlab{a}})Kirschstein, Giebenhain, and
  Nie{\ss}ner]{kirschstein2024diffusionavatars}
Tobias Kirschstein, Simon Giebenhain, and Matthias Nie{\ss}ner.
\newblock Diffusionavatars: Deferred diffusion for high-fidelity 3d head
  avatars.
\newblock In \emph{Proceedings of the IEEE/CVF Conference on Computer Vision
  and Pattern Recognition}, pages 5481--5492, 2024{\natexlab{a}}.

\bibitem[Kirschstein et~al.(2024{\natexlab{b}})Kirschstein, Giebenhain, Tang,
  Georgopoulos, and Nie{\ss}ner]{kirschstein2024gghead}
Tobias Kirschstein, Simon Giebenhain, Jiapeng Tang, Markos Georgopoulos, and
  Matthias Nie{\ss}ner.
\newblock Gghead: Fast and generalizable 3d gaussian heads.
\newblock \emph{arXiv preprint arXiv:2406.09377}, 2024{\natexlab{b}}.

\bibitem[Lee et~al.(2022)Lee, Choi, Go, Lee, Cho, and Kim]{lee2022expgan}
Yeonkyeong Lee, Taeho Choi, Hyunsung Go, Hyunjoon Lee, Sunghyun Cho, and Junho
  Kim.
\newblock Exp-gan: 3d-aware facial image generation with expression control.
\newblock In \emph{Proceedings of the Asian Conference on Computer Vision},
  pages 3812--3827, 2022.

\bibitem[Li et~al.(2024)Li, Cao, Schwartz, Khirodkar, Richardt, Simon, Sheikh,
  and Saito]{li2024uravatar}
Junxuan Li, Chen Cao, Gabriel Schwartz, Rawal Khirodkar, Christian Richardt,
  Tomas Simon, Yaser Sheikh, and Shunsuke Saito.
\newblock Uravatar: Universal relightable gaussian codec avatars.
\newblock \emph{arXiv preprint arXiv:2410.24223}, 2024.

\bibitem[Li et~al.(2017)Li, Bolkart, Black, Li, and Romero]{li2017flame}
Tianye Li, Timo Bolkart, Michael~J Black, Hao Li, and Javier Romero.
\newblock Learning a model of facial shape and expression from 4d scans.
\newblock \emph{ACM Trans. Graph.}, 36\penalty0 (6):\penalty0 194--1, 2017.

\bibitem[Lin et~al.(2020)Lin, Ryabtsev, Sengupta, Curless, Seitz, and
  Kemelmacher-Shlizerman]{BGMv2}
Shanchuan Lin, Andrey Ryabtsev, Soumyadip Sengupta, Brian Curless, Steve Seitz,
  and Ira Kemelmacher-Shlizerman.
\newblock Real-time high-resolution background matting.
\newblock \emph{arXiv}, pages arXiv--2012, 2020.

\bibitem[Liu et~al.(2025)Liu, Wang, Hu, Shen, Ye, Zang, Cao, Li, and
  Liu]{liu2025mvsgaussian}
Tianqi Liu, Guangcong Wang, Shoukang Hu, Liao Shen, Xinyi Ye, Yuhang Zang,
  Zhiguo Cao, Wei Li, and Ziwei Liu.
\newblock Mvsgaussian: Fast generalizable gaussian splatting reconstruction
  from multi-view stereo.
\newblock In \emph{European Conference on Computer Vision}, pages 37--53.
  Springer, 2025.

\bibitem[Liu et~al.(2022)Liu, Shu, Li, Lin, Zhang, and Kung]{liu20223dfmgan}
Yuchen Liu, Zhixin Shu, Yijun Li, Zhe Lin, Richard Zhang, and Sun-Yuan Kung.
\newblock 3d-fm gan: Towards 3d-controllable face manipulation.
\newblock In \emph{European conference on computer vision}, pages 107--125.
  Springer, 2022.

\bibitem[Lombardi et~al.(2021)Lombardi, Simon, Schwartz, Zollhoefer, Sheikh,
  and Saragih]{lombardi2021mvp}
Stephen Lombardi, Tomas Simon, Gabriel Schwartz, Michael Zollhoefer, Yaser
  Sheikh, and Jason Saragih.
\newblock Mixture of volumetric primitives for efficient neural rendering.
\newblock \emph{ACM Transactions on Graphics (ToG)}, 40\penalty0 (4):\penalty0
  1--13, 2021.

\bibitem[Lyu et~al.(2024)Lyu, Zhou, Yang, and Shu]{lyu2024facelift}
Weijie Lyu, Yi Zhou, Ming-Hsuan Yang, and Zhixin Shu.
\newblock Facelift: Single image to 3d head with view generation and gs-lrm,
  2024.

\bibitem[Martinez et~al.(2024)Martinez, Kim, Romero, Bagautdinov, Saito, Yu,
  Anderson, Zollhöfer, Wang, Bai, Li, Wei, Joshi, Borsos, Simon, Saragih,
  Theodosis, Greene, Josyula, Maeta, Jewett, Venshtain, Heilman, Chen, Fu,
  Elshaer, Du, Wu, Chen, Kang, Wu, Emad, Longay, Brewer, Shah, Booth, Koska,
  Haidle, Andromalos, Hsu, Dauer, Selednik, Godisart, Ardisson, Cipperly,
  Humberston, Farr, Hansen, Guo, Braun, Krenn, Wen, Evans, Fadeeva, Stewart,
  Schwartz, Gupta, Moon, Guo, Dong, Xu, Shiratori, Prada, Pires, Peng,
  Buffalini, Trimble, McPhail, Schoeller, and Sheikh]{ava256}
Julieta Martinez, Emily Kim, Javier Romero, Timur Bagautdinov, Shunsuke Saito,
  Shoou-I Yu, Stuart Anderson, Michael Zollhöfer, Te-Li Wang, Shaojie Bai,
  Chenghui Li, Shih-En Wei, Rohan Joshi, Wyatt Borsos, Tomas Simon, Jason
  Saragih, Paul Theodosis, Alexander Greene, Anjani Josyula, Silvio~Mano Maeta,
  Andrew~I. Jewett, Simon Venshtain, Christopher Heilman, Yueh-Tung Chen, Sidi
  Fu, Mohamed Ezzeldin~A. Elshaer, Tingfang Du, Longhua Wu, Shen-Chi Chen, Kai
  Kang, Michael Wu, Youssef Emad, Steven Longay, Ashley Brewer, Hitesh Shah,
  James Booth, Taylor Koska, Kayla Haidle, Matt Andromalos, Joanna Hsu, Thomas
  Dauer, Peter Selednik, Tim Godisart, Scott Ardisson, Matthew Cipperly, Ben
  Humberston, Lon Farr, Bob Hansen, Peihong Guo, Dave Braun, Steven Krenn, He
  Wen, Lucas Evans, Natalia Fadeeva, Matthew Stewart, Gabriel Schwartz, Divam
  Gupta, Gyeongsik Moon, Kaiwen Guo, Yuan Dong, Yichen Xu, Takaaki Shiratori,
  Fabian Prada, Bernardo~R. Pires, Bo Peng, Julia Buffalini, Autumn Trimble,
  Kevyn McPhail, Melissa Schoeller, and Yaser Sheikh.
\newblock {Codec Avatar Studio: Paired Human Captures for Complete, Driveable,
  and Generalizable Avatars}.
\newblock \emph{NeurIPS Track on Datasets and Benchmarks}, 2024.

\bibitem[Mildenhall et~al.(2021)Mildenhall, Srinivasan, Tancik, Barron,
  Ramamoorthi, and Ng]{mildenhall2021nerf}
Ben Mildenhall, Pratul~P Srinivasan, Matthew Tancik, Jonathan~T Barron, Ravi
  Ramamoorthi, and Ren Ng.
\newblock Nerf: Representing scenes as neural radiance fields for view
  synthesis.
\newblock \emph{Communications of the ACM}, 65\penalty0 (1):\penalty0 99--106,
  2021.

\bibitem[Pan et~al.(2024)Pan, Zhuo, Piao, Luo, Cheng, Wang, Fan, Liu, Yang,
  Dai, et~al.]{pan2024renderme360}
Dongwei Pan, Long Zhuo, Jingtan Piao, Huiwen Luo, Wei Cheng, Yuxin Wang, Siming
  Fan, Shengqi Liu, Lei Yang, Bo Dai, et~al.
\newblock Renderme-360: a large digital asset library and benchmarks towards
  high-fidelity head avatars.
\newblock \emph{Advances in Neural Information Processing Systems}, 36, 2024.

\bibitem[Qian et~al.(2024)Qian, Kirschstein, Schoneveld, Davoli, Giebenhain,
  and Nie{\ss}ner]{qian2024gaussianavatars}
Shenhan Qian, Tobias Kirschstein, Liam Schoneveld, Davide Davoli, Simon
  Giebenhain, and Matthias Nie{\ss}ner.
\newblock Gaussianavatars: Photorealistic head avatars with rigged 3d
  gaussians.
\newblock In \emph{Proceedings of the IEEE/CVF Conference on Computer Vision
  and Pattern Recognition}, pages 20299--20309, 2024.

\bibitem[Ren et~al.(2024)Ren, Xie, Mirzaei, Liang, Zeng, Kreis, Liu, Torralba,
  Fidler, Kim, et~al.]{ren2024l4gm}
Jiawei Ren, Kevin Xie, Ashkan Mirzaei, Hanxue Liang, Xiaohui Zeng, Karsten
  Kreis, Ziwei Liu, Antonio Torralba, Sanja Fidler, Seung~Wook Kim, et~al.
\newblock L4gm: Large 4d gaussian reconstruction model.
\newblock \emph{arXiv preprint arXiv:2406.10324}, 2024.

\bibitem[Saito et~al.(2024)Saito, Schwartz, Simon, Li, and Nam]{saito2024rgca}
Shunsuke Saito, Gabriel Schwartz, Tomas Simon, Junxuan Li, and Giljoo Nam.
\newblock Relightable gaussian codec avatars.
\newblock In \emph{Proceedings of the IEEE/CVF Conference on Computer Vision
  and Pattern Recognition}, pages 130--141, 2024.

\bibitem[Sun et~al.(2022)Sun, Wang, Shi, Wang, Wang, and Liu]{sun2022ide3d}
Jingxiang Sun, Xuan Wang, Yichun Shi, Lizhen Wang, Jue Wang, and Yebin Liu.
\newblock Ide-3d: Interactive disentangled editing for high-resolution 3d-aware
  portrait synthesis.
\newblock \emph{ACM Transactions on Graphics (ToG)}, 41\penalty0 (6):\penalty0
  1--10, 2022.

\bibitem[Sun et~al.(2023)Sun, Wang, Wang, Li, Zhang, Zhang, and
  Liu]{sun2023next3d}
Jingxiang Sun, Xuan Wang, Lizhen Wang, Xiaoyu Li, Yong Zhang, Hongwen Zhang,
  and Yebin Liu.
\newblock Next3d: Generative neural texture rasterization for 3d-aware head
  avatars.
\newblock In \emph{Proceedings of the IEEE/CVF conference on computer vision
  and pattern recognition}, pages 20991--21002, 2023.

\bibitem[Tang et~al.(2023)Tang, Zhang, Yang, Zhang, Chen, Ma, and
  Wen]{tang20233dfaceshop}
Junshu Tang, Bo Zhang, Binxin Yang, Ting Zhang, Dong Chen, Lizhuang Ma, and
  Fang Wen.
\newblock 3dfaceshop: Explicitly controllable 3d-aware portrait generation.
\newblock \emph{IEEE transactions on visualization and computer graphics},
  30\penalty0 (9):\penalty0 6020--6037, 2023.

\bibitem[Tang et~al.(2024)Tang, Davoli, Kirschstein, Schoneveld, and
  Niessner]{tang2024gaf}
Jiapeng Tang, Davide Davoli, Tobias Kirschstein, Liam Schoneveld, and Matthias
  Niessner.
\newblock Gaf: Gaussian avatar reconstruction from monocular videos via
  multi-view diffusion.
\newblock \emph{arXiv preprint arXiv:2412.10209}, 2024.

\bibitem[Tang et~al.(2025)Tang, Chen, Chen, Wang, Zeng, and Liu]{tang2025lgm}
Jiaxiang Tang, Zhaoxi Chen, Xiaokang Chen, Tengfei Wang, Gang Zeng, and Ziwei
  Liu.
\newblock Lgm: Large multi-view gaussian model for high-resolution 3d content
  creation.
\newblock In \emph{European Conference on Computer Vision}, pages 1--18.
  Springer, 2025.

\bibitem[Taubner et~al.(2024)Taubner, Zhang, Tuli, and
  Lindell]{taubner2024cap4d}
Felix Taubner, Ruihang Zhang, Mathieu Tuli, and David~B Lindell.
\newblock Cap4d: Creating animatable 4d portrait avatars with morphable
  multi-view diffusion models.
\newblock \emph{arXiv preprint arXiv:2412.12093}, 2024.

\bibitem[Teotia et~al.(2024{\natexlab{a}})Teotia, Kim, Garrido, Habermann,
  Elgharib, and Theobalt]{teotia2024gaussianheads}
Kartik Teotia, Hyeongwoo Kim, Pablo Garrido, Marc Habermann, Mohamed Elgharib,
  and Christian Theobalt.
\newblock Gaussianheads: End-to-end learning of drivable gaussian head avatars
  from coarse-to-fine representations.
\newblock \emph{ACM Transactions on Graphics (TOG)}, 43\penalty0 (6):\penalty0
  1--12, 2024{\natexlab{a}}.

\bibitem[Teotia et~al.(2024{\natexlab{b}})Teotia, Pan, Kim, Garrido, Elgharib,
  and Theobalt]{teotia2024hq3davatar}
Kartik Teotia, Xingang Pan, Hyeongwoo Kim, Pablo Garrido, Mohamed Elgharib, and
  Christian Theobalt.
\newblock Hq3davatar: High-quality implicit 3d head avatar.
\newblock \emph{ACM Transactions on Graphics}, 43\penalty0 (3):\penalty0 1--24,
  2024{\natexlab{b}}.

\bibitem[Tewari et~al.(2020)Tewari, Elgharib, Bharaj, Bernard, Seidel,
  P{\'e}rez, Zollhofer, and Theobalt]{tewari2020stylerig}
Ayush Tewari, Mohamed Elgharib, Gaurav Bharaj, Florian Bernard, Hans-Peter
  Seidel, Patrick P{\'e}rez, Michael Zollhofer, and Christian Theobalt.
\newblock Stylerig: Rigging stylegan for 3d control over portrait images.
\newblock In \emph{Proceedings of the IEEE/CVF conference on computer vision
  and pattern recognition}, pages 6142--6151, 2020.

\bibitem[Tran et~al.(2024{\natexlab{a}})Tran, Zakharov, Ho, Hu, Karmanov,
  Agarwal, Goldwhite, Venegas, Tran, and Li]{tran2024voodooxp}
Phong Tran, Egor Zakharov, Long-Nhat Ho, Liwen Hu, Adilbek Karmanov, Aviral
  Agarwal, McLean Goldwhite, Ariana~Bermudez Venegas, Anh~Tuan Tran, and Hao
  Li.
\newblock Voodoo xp: Expressive one-shot head reenactment for vr telepresence.
\newblock \emph{arXiv preprint arXiv:2405.16204}, 2024{\natexlab{a}}.

\bibitem[Tran et~al.(2024{\natexlab{b}})Tran, Zakharov, Ho, Tran, Hu, and
  Li]{tran2024voodoo3d}
Phong Tran, Egor Zakharov, Long-Nhat Ho, Anh~Tuan Tran, Liwen Hu, and Hao Li.
\newblock Voodoo 3d: Volumetric portrait disentanglement for one-shot 3d head
  reenactment.
\newblock In \emph{Proceedings of the IEEE/CVF Conference on Computer Vision
  and Pattern Recognition}, pages 10336--10348, 2024{\natexlab{b}}.

\bibitem[Trevithick et~al.(2023)Trevithick, Chan, Takikawa, Iqbal, Mello,
  Chandraker, Ramamoorthi, and and]{Trevithick2023wysiwyg}
Alex Trevithick, Matthew Chan, Towaki Takikawa, Umar Iqbal, Shalini~De Mello,
  Manmohan Chandraker, Ravi Ramamoorthi, and Koki~Nagano and.
\newblock Rendering every pixel for high-fidelity geometry in 3d gans.
\newblock In \emph{arXiv}, 2023.

\bibitem[Wang et~al.(2023)Wang, Tan, Bi, Xu, Luan, Sunkavalli, Wang, Xu, and
  Zhang]{wang2023pflrm}
Peng Wang, Hao Tan, Sai Bi, Yinghao Xu, Fujun Luan, Kalyan Sunkavalli, Wenping
  Wang, Zexiang Xu, and Kai Zhang.
\newblock Pf-lrm: Pose-free large reconstruction model for joint pose and shape
  prediction.
\newblock \emph{arXiv preprint arXiv:2311.12024}, 2023.

\bibitem[Wang et~al.(2024)Wang, Leroy, Cabon, Chidlovskii, and
  Revaud]{wang2024dust3r}
Shuzhe Wang, Vincent Leroy, Yohann Cabon, Boris Chidlovskii, and Jerome Revaud.
\newblock Dust3r: Geometric 3d vision made easy.
\newblock In \emph{Proceedings of the IEEE/CVF Conference on Computer Vision
  and Pattern Recognition}, pages 20697--20709, 2024.

\bibitem[Wang et~al.(2004)Wang, Bovik, Sheikh, and Simoncelli]{wang2004ssim}
Zhou Wang, Alan~C Bovik, Hamid~R Sheikh, and Eero~P Simoncelli.
\newblock Image quality assessment: from error visibility to structural
  similarity.
\newblock \emph{IEEE transactions on image processing}, 13\penalty0
  (4):\penalty0 600--612, 2004.

\bibitem[Xiang et~al.(2024)Xiang, Gao, Guo, and Zhang]{xiang2024flashavatar}
Jun Xiang, Xuan Gao, Yudong Guo, and Juyong Zhang.
\newblock Flashavatar: High-fidelity head avatar with efficient gaussian
  embedding.
\newblock In \emph{Proceedings of the IEEE/CVF Conference on Computer Vision
  and Pattern Recognition}, pages 1802--1812, 2024.

\bibitem[Xie et~al.(2022)Xie, Wang, Zhang, Dong, and Shan]{xie2022vfhq}
Liangbin Xie, Xintao Wang, Honglun Zhang, Chao Dong, and Ying Shan.
\newblock Vfhq: A high-quality dataset and benchmark for video face
  super-resolution.
\newblock In \emph{The IEEE Conference on Computer Vision and Pattern
  Recognition Workshops (CVPRW)}, 2022.

\bibitem[Xu et~al.(2023)Xu, Wang, Zhao, Zhang, and Liu]{xu2023avatarmav}
Yuelang Xu, Lizhen Wang, Xiaochen Zhao, Hongwen Zhang, and Yebin Liu.
\newblock Avatarmav: Fast 3d head avatar reconstruction using motion-aware
  neural voxels.
\newblock In \emph{ACM SIGGRAPH 2023 Conference Proceedings}, pages 1--10,
  2023.

\bibitem[Xu et~al.(2024{\natexlab{a}})Xu, Chen, Li, Zhang, Wang, Zheng, and
  Liu]{xu2023gha}
Yuelang Xu, Benwang Chen, Zhe Li, Hongwen Zhang, Lizhen Wang, Zerong Zheng, and
  Yebin Liu.
\newblock Gaussian head avatar: Ultra high-fidelity head avatar via dynamic
  gaussians.
\newblock In \emph{Proceedings of the IEEE/CVF Conference on Computer Vision
  and Pattern Recognition (CVPR)}, 2024{\natexlab{a}}.

\bibitem[Xu et~al.(2024{\natexlab{b}})Xu, Shi, Yifan, Chen, Yang, Peng, Shen,
  and Wetzstein]{xu2024grm}
Yinghao Xu, Zifan Shi, Wang Yifan, Hansheng Chen, Ceyuan Yang, Sida Peng, Yujun
  Shen, and Gordon Wetzstein.
\newblock Grm: Large gaussian reconstruction model for efficient 3d
  reconstruction and generation.
\newblock \emph{arXiv preprint arXiv:2403.14621}, 2024{\natexlab{b}}.

\bibitem[Xu et~al.(2025)Xu, Wang, Zheng, Su, and Liu]{xu2025gphm}
Yuelang Xu, Lizhen Wang, Zerong Zheng, Zhaoqi Su, and Yebin Liu.
\newblock 3d gaussian parametric head model.
\newblock In \emph{European Conference on Computer Vision}, pages 129--147.
  Springer, 2025.

\bibitem[Yang et~al.(2020)Yang, Zhu, Wang, Huang, Shen, Yang, and
  Cao]{yang2020facescape}
Haotian Yang, Hao Zhu, Yanru Wang, Mingkai Huang, Qiu Shen, Ruigang Yang, and
  Xun Cao.
\newblock Facescape: a large-scale high quality 3d face dataset and detailed
  riggable 3d face prediction.
\newblock In \emph{Proceedings of the ieee/cvf conference on computer vision
  and pattern recognition}, pages 601--610, 2020.

\bibitem[Yang et~al.(2024)Yang, Zheng, Ma, Lai, Wan, and Huang]{yang2024vrmm}
Haotian Yang, Mingwu Zheng, Chongyang Ma, Yu-Kun Lai, Pengfei Wan, and Haibin
  Huang.
\newblock Vrmm: A volumetric relightable morphable head model.
\newblock In \emph{ACM SIGGRAPH 2024 Conference Papers}, pages 1--11, 2024.

\bibitem[Ye et~al.(2023)Ye, Zhang, Liu, Han, and Yang]{ye2023ipadapter}
Hu Ye, Jun Zhang, Sibo Liu, Xiao Han, and Wei Yang.
\newblock Ip-adapter: Text compatible image prompt adapter for text-to-image
  diffusion models.
\newblock \emph{arXiv preprint arXiv:2308.06721}, 2023.

\bibitem[Yu et~al.(2024)Yu, Bai, Meka, Tan, Xu, Pandey, Fanello, Park, and
  Zhang]{yu2024one2avatar}
Zhixuan Yu, Ziqian Bai, Abhimitra Meka, Feitong Tan, Qiangeng Xu, Rohit Pandey,
  Sean Fanello, Hyun~Soo Park, and Yinda Zhang.
\newblock One2avatar: Generative implicit head avatar for few-shot user
  adaptation.
\newblock \emph{arXiv preprint arXiv:2402.11909}, 2024.

\bibitem[Yuan et~al.(2023)Yuan, Zhu, Li, Liu, and Yuan]{yuan2023goae}
Ziyang Yuan, Yiming Zhu, Yu Li, Hongyu Liu, and Chun Yuan.
\newblock Make encoder great again in 3d gan inversion through geometry and
  occlusion-aware encoding.
\newblock In \emph{Proceedings of the IEEE/CVF International Conference on
  Computer Vision}, pages 2437--2447, 2023.

\bibitem[Zhang et~al.(2025)Zhang, Bi, Tan, Xiangli, Zhao, Sunkavalli, and
  Xu]{zhang2025gslrm}
Kai Zhang, Sai Bi, Hao Tan, Yuanbo Xiangli, Nanxuan Zhao, Kalyan Sunkavalli,
  and Zexiang Xu.
\newblock Gs-lrm: Large reconstruction model for 3d gaussian splatting.
\newblock In \emph{European Conference on Computer Vision}, pages 1--19.
  Springer, 2025.

\bibitem[Zhang et~al.(2018)Zhang, Isola, Efros, Shechtman, and
  Wang]{zhang2018lpips}
Richard Zhang, Phillip Isola, Alexei~A Efros, Eli Shechtman, and Oliver Wang.
\newblock The unreasonable effectiveness of deep features as a perceptual
  metric.
\newblock In \emph{Proceedings of the IEEE conference on computer vision and
  pattern recognition}, pages 586--595, 2018.

\bibitem[Zhao et~al.(2024)Zhao, Sun, Wang, Suo, and Liu]{zhao2024invertavatar}
Xiaochen Zhao, Jingxiang Sun, Lizhen Wang, Jinli Suo, and Yebin Liu.
\newblock Invertavatar: Incremental gan inversion for generalized head avatars.
\newblock In \emph{ACM SIGGRAPH 2024 Conference Papers}, pages 1--10, 2024.

\bibitem[Zheng et~al.(2024)Zheng, Wen, Li, Zhang, Su, Chang, Zhao, Lv, Zhang,
  Zhang, et~al.]{zheng2024headgap}
Xiaozheng Zheng, Chao Wen, Zhaohu Li, Weiyi Zhang, Zhuo Su, Xu Chang, Yang
  Zhao, Zheng Lv, Xiaoyuan Zhang, Yongjie Zhang, et~al.
\newblock Headgap: Few-shot 3d head avatar via generalizable gaussian priors.
\newblock \emph{arXiv preprint arXiv:2408.06019}, 2024.

\bibitem[Zheng et~al.(2022)Zheng, Abrevaya, B{\"u}hler, Chen, Black, and
  Hilliges]{zheng2022imavatar}
Yufeng Zheng, Victoria~Fern{\'a}ndez Abrevaya, Marcel~C B{\"u}hler, Xu Chen,
  Michael~J Black, and Otmar Hilliges.
\newblock Im avatar: Implicit morphable head avatars from videos.
\newblock In \emph{Proceedings of the IEEE/CVF conference on computer vision
  and pattern recognition}, pages 13545--13555, 2022.

\bibitem[Zheng et~al.(2023)Zheng, Yifan, Wetzstein, Black, and
  Hilliges]{zheng2023pointavatar}
Yufeng Zheng, Wang Yifan, Gordon Wetzstein, Michael~J Black, and Otmar
  Hilliges.
\newblock Pointavatar: Deformable point-based head avatars from videos.
\newblock In \emph{Proceedings of the IEEE/CVF conference on computer vision
  and pattern recognition}, pages 21057--21067, 2023.

\bibitem[Zielonka et~al.(2022)Zielonka, Bolkart, and Thies]{MICA:ECCV2022}
Wojciech Zielonka, Timo Bolkart, and Justus Thies.
\newblock Towards metrical reconstruction of human faces.
\newblock In \emph{European conference on computer vision}, pages 250--269.
  Springer, 2022.

\bibitem[Zielonka et~al.(2023{\natexlab{a}})Zielonka, Bagautdinov, Saito,
  Zollh{\"o}fer, Thies, and Romero]{zielonka2023d3ga}
Wojciech Zielonka, Timur Bagautdinov, Shunsuke Saito, Michael Zollh{\"o}fer,
  Justus Thies, and Javier Romero.
\newblock Drivable 3d gaussian avatars.
\newblock \emph{arXiv preprint arXiv:2311.08581}, 2023{\natexlab{a}}.

\bibitem[Zielonka et~al.(2023{\natexlab{b}})Zielonka, Bolkart, and
  Thies]{zielonka2023insta}
Wojciech Zielonka, Timo Bolkart, and Justus Thies.
\newblock Instant volumetric head avatars.
\newblock In \emph{Proceedings of the IEEE/CVF conference on computer vision
  and pattern recognition}, pages 4574--4584, 2023{\natexlab{b}}.

\end{thebibliography}
}

\clearpage
\setcounter{page}{1}
\setcounter{table}{2}
\setcounter{figure}{8}
\maketitlesupplementary

\appendix

\begin{table}[tb]
    \setlength{\tabcolsep}{2pt}
    \centering
    \begin{tabular}{lclrrrrr}
        \toprule
            &\scriptsize{Inputs} && \scriptsize{PSNR}$\uparrow$ & \scriptsize{SSIM}$\uparrow$ & \scriptsize{LPIPS}$\downarrow$ & \scriptsize{AKD}$\downarrow$ & \scriptsize{CSIM}$\uparrow$ \\

        \midrule
          \multirow{4}{*}{\rotatebox[origin=c]{90}{\parbox[c]{1.2cm}{\centering \footnotesize Ava256}}}
                & 1 & GAGAvatar
                    & 18.1          & 0.66          & 0.37          & 7.0          & \textbf{0.45}\\
                & 1 & GAGAvatar\textsuperscript{\textdagger}
                    & \textbf{19.6}          & 0.68          & \textbf{0.31}          & \textbf{5.6}	       & 0.33\\
                & 1 & Ours\textsuperscript{3DGAN}
                    & 19.1          & 0.68          & 0.39          & 6.0          & 0.37\\
                & 1 & Ours\textsuperscript{1}
                    & \textbf{19.6}          & \textbf{0.70}          & 0.46          & 5.7          & 0.40\\
        \midrule
            \multirow{11}{*}{\rotatebox[origin=c]{90}{\parbox[c]{1.5cm}{\centering \footnotesize NeRSemble}}}
                & Video & FlashAvatar
                    & 15.0& 0.42 & 0.61 & 8.8 & 0.17 \\
                & 4 & Ours 
                    & 20.5 & 0.75 & 0.33 & \textbf{3.7} & 0.50 \\
                & 4 & Ours\textsuperscript{+ 984} 
                    & \textbf{21.6} & \textbf{0.77} & \textbf{0.29} & 3.8 & \textbf{0.76} \\
            \cmidrule(l){2-8}
                & 1 & HeadNeRF
                    & 9.7           & 0.69          & 0.48          & 5.2          & 0.18\\
                & 1 & Portrait4D-v2
                    & 17.5          & 0.58          & 0.36          & 5.4          & 0.41\\ %
                & 1 & GAGAvatar
                    & 18.7          & 0.70          & 0.35          & 5.4          & 0.44\\ %
                & 1 & GAGAvatar\textsuperscript{\textdagger}
                    & 18.9          & 0.70          & \textbf{0.32} & 4.8          & 0.24\\ %
                & 1 & Ours\textsuperscript{3DGAN}
                    & 19.5          & 0.71          & 0.38          & \textbf{4.3} & 0.31\\
                & 1 & Ours\textsuperscript{3DGAN + 984} 
                    & 19.4 & 0.71 & 0.38 & 4.6 & \textbf{0.46} \\
                & 1 & Ours\textsuperscript{1}
                    & \textbf{19.8} & \textbf{0.73} & 0.38          & 4.4          & 0.30\\
        \bottomrule
        \multicolumn{3}{l}{\scriptsize{\textsuperscript{\textdagger}re-trained on Ava256}}
        & \multicolumn{4}{l}{\scriptsize{\textsuperscript{1}trained on 1 input view}}
    \end{tabular}
    \caption{\textbf{Quantitative Comparison on 3D head avatar creation from various.} Our approach performs competitively when only a single input image is available, despite not being designed for a single image use-case. It also performs much better compared to monocular methods that receive a full video as input. Ours\textsuperscript{3DGAN} denotes a 4-shot model where the 4 required images are obtained from the single input via 3D lifting with a 3D GAN. Ours\textsuperscript{+ 984} denotes a model that is additionally fine-tuned on 984 neutral identities for better identity preservation (CSIM metric). }
    \label{tab:single_image_comparison}
\end{table}

\section{Additional Results}
\subsection{Avat3rs from Phone Captures and Accessories}
\begin{figure*}
    \centering
    \includegraphics[width=\linewidth]{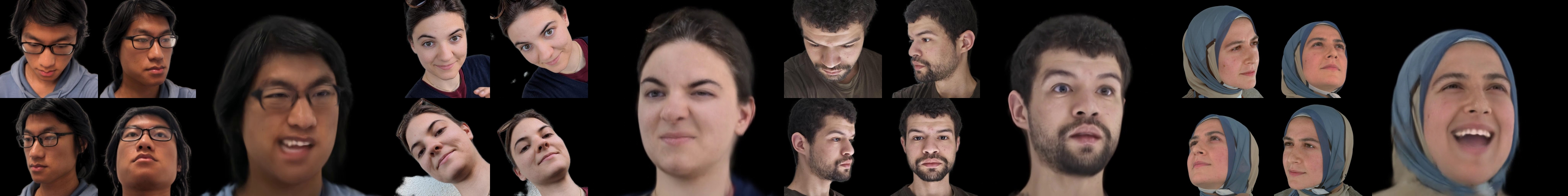}
    \caption{\textbf{More Avat3r reconstruction results} from phone scans with different phone cameras and persons with challenging clothing.}
    \label{fig:phone_scans}
\end{figure*}

~\cref{fig:phone_scans} showcases additional phone scans captured by users on their own devices, including challenging cases with glasses, rotated inputs, and a NeRSemble subject wearing a headscarf. Despite not being trained on accessories, the model handles glasses reasonably well and reconstructs the headscarf with high fidelity.

\begin{figure}[tb]
    \setlength{\tabcolsep}{0pt}
    \centering
    \vspace{0.2cm}
    \begin{tabularx}{\linewidth}{P{0.25\linewidth}P{0.25\linewidth}P{0.25\linewidth}P{0.25\linewidth}}
         \scriptsize{FlashAvatar frontal} & \scriptsize{FlashAvatar} & \scriptsize{Ours} & \scriptsize{GT}\\
    \end{tabularx}
    
    \includegraphics[width=\linewidth]{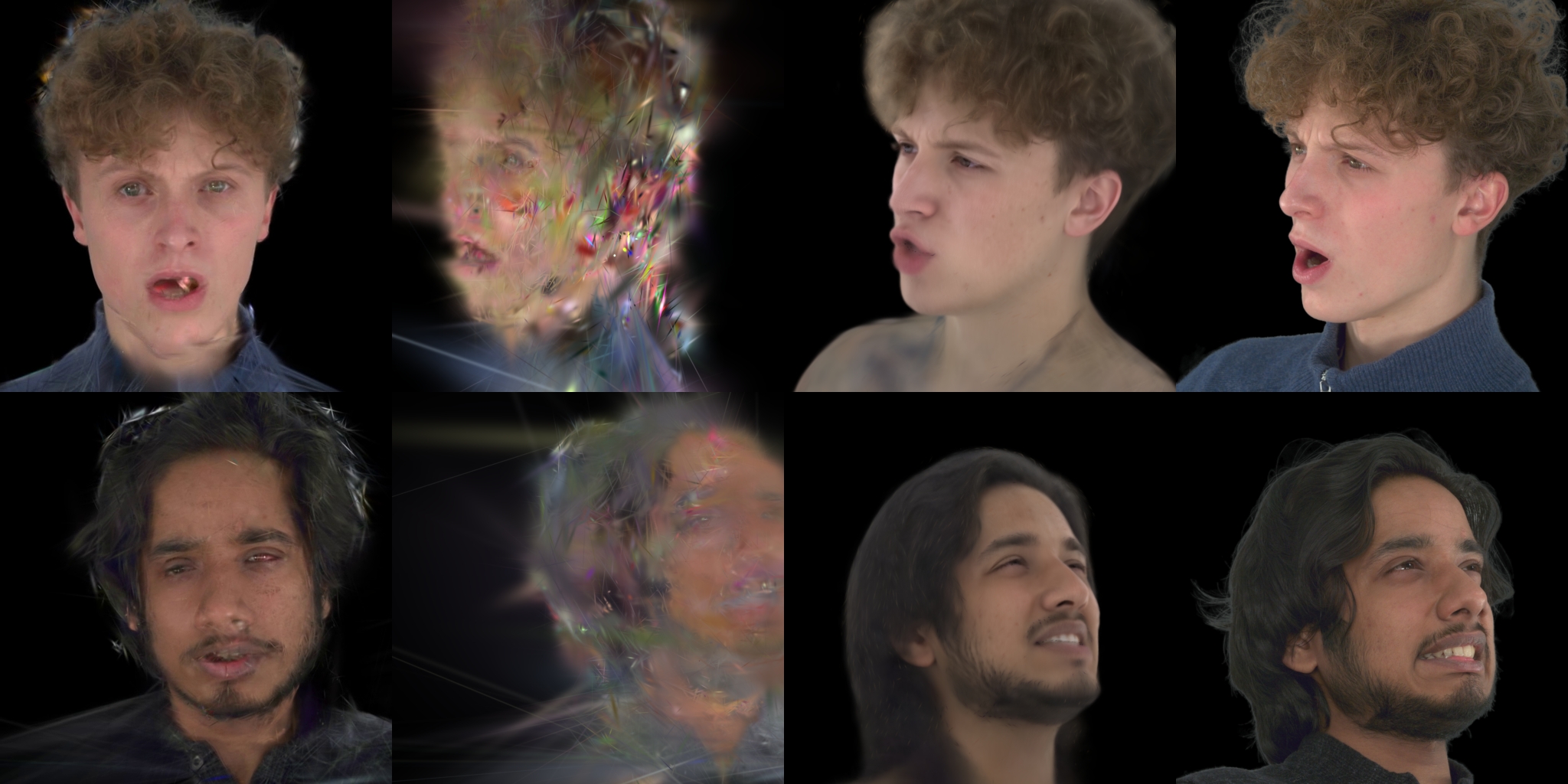}
    
    \vspace{-0.3cm}
    \caption{\textbf{Comparison with FlashAvatar} on NeRSemble.}
    \vspace{-0.2cm}
    \label{fig:iccv_rebuttal_flash_avatar_comparison}

\end{figure}

\begin{figure*}
    \centering
    \includegraphics[width=\linewidth]{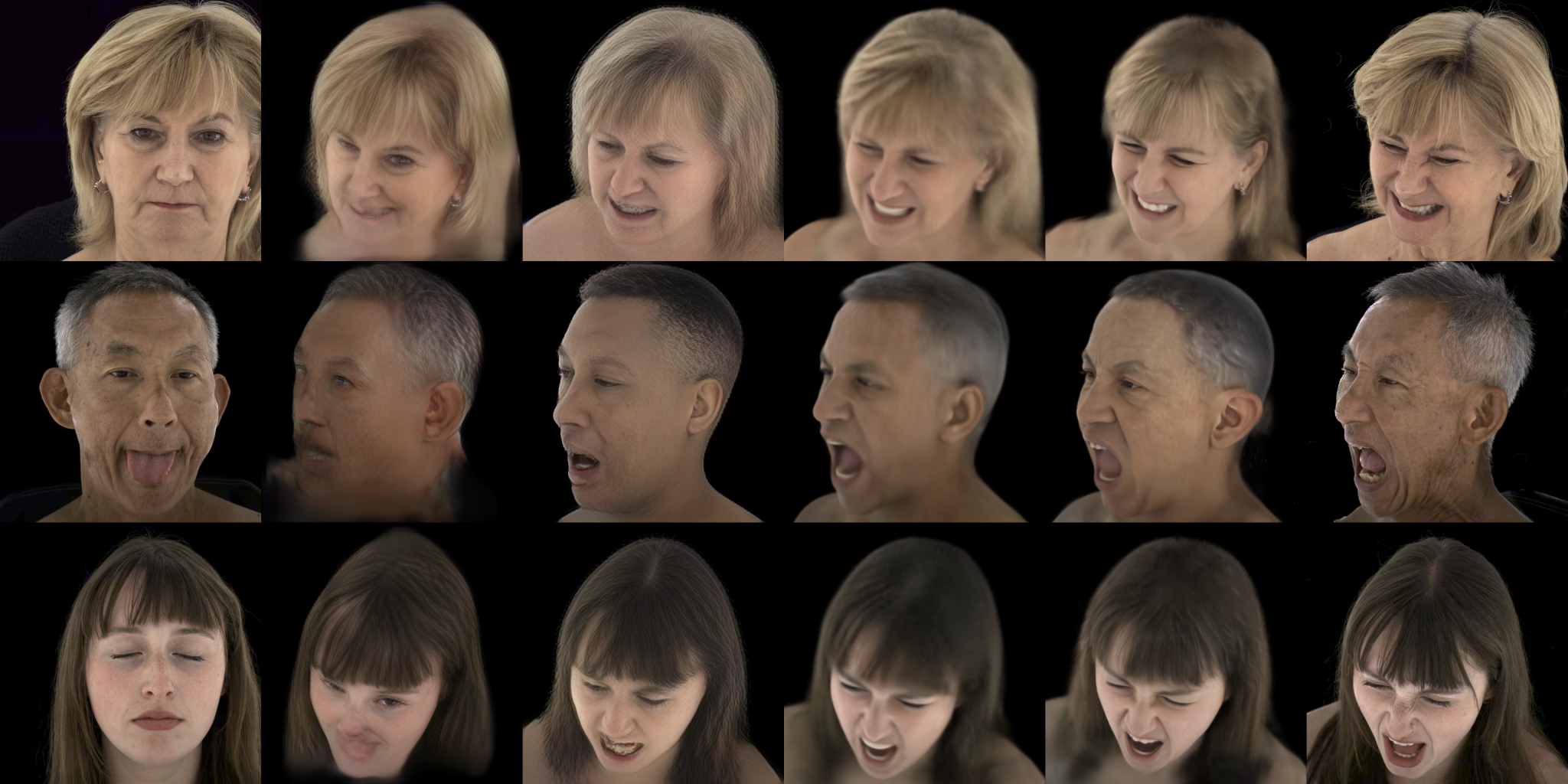}
    \begin{tabularx}{\linewidth}{YYYYYY}
        Input & GAGavatar & GAGavatar\textsuperscript{\textdagger} & Ours\textsuperscript{1} & Ours\textsuperscript{3DGAN} & GT
    \end{tabularx}
    \caption{\textbf{Single-image comparison on Ava256.} We compare Avat3r with the recent 3D-aware portrait animation method GAGAvatar~\cite{chu2024gagavatar} in a self-reenactment scenario on hold-out persons from the Ava256 dataset. GAGavatar\textsuperscript{\textdagger} denotes a version of the baseline that we trained on the Ava256 dataset. Ours\textsuperscript{1} is a version of our model that was trained on only 1 input image (see~\cref{sec:data_efficiency}). Our method with 3D lifting (Ours\textsuperscript{3DGAN}) shows better rendering quality than the baselines, especially for extreme expressions and viewing angles.}
    \label{fig:4_single_shot_comparison_ava}
\end{figure*}

\begin{figure*}
    \centering
    \includegraphics[width=\linewidth]{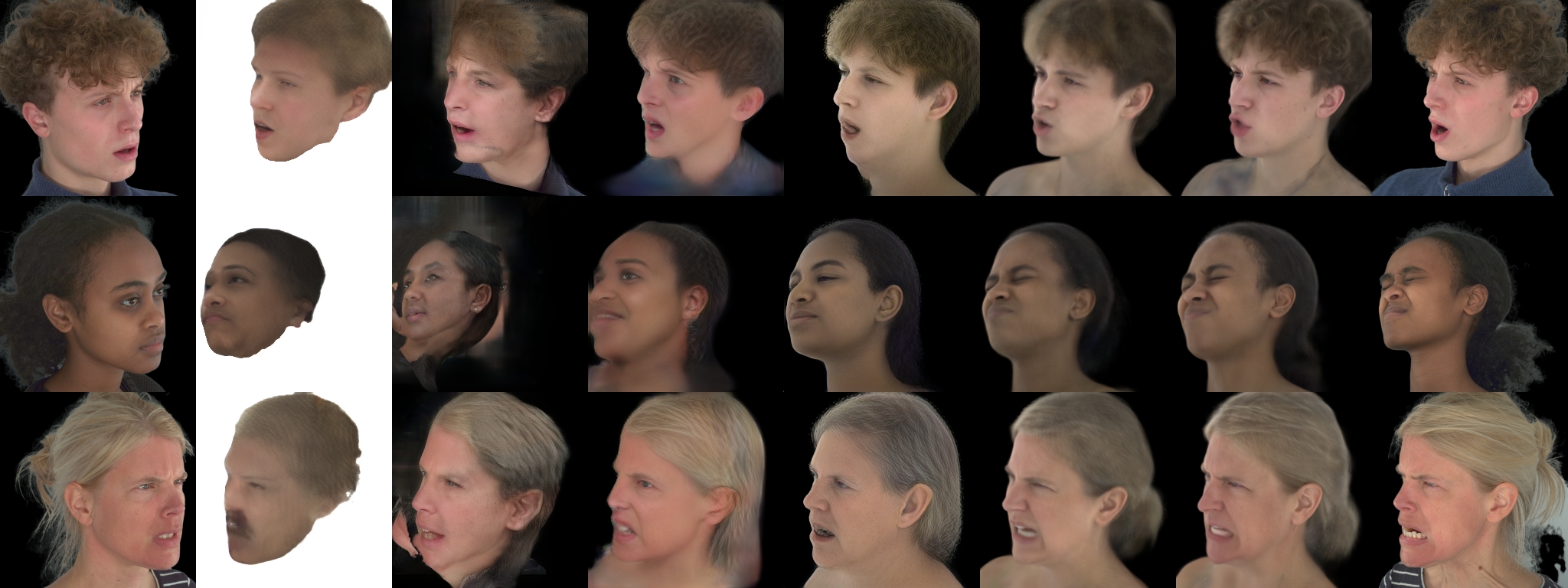}
    \begin{tabularx}{\linewidth}{YYYYYYYY}
        Input & HeadNeRF & Portrait4Dv2 & GAGavatar & GAGavatar\textsuperscript{\textdagger} & Ours\textsuperscript{1} & Ours\textsuperscript{3DGAN} & GT
    \end{tabularx}
    \caption{\textbf{Single-image comparison on NeRSemble.} We compare Avat3r with the recent 3D-aware portrait animation methods GAGAvatar~\cite{chu2024gagavatar} (GAGavatar\textsuperscript{\textdagger} denotes a version of the baseline that we trained on the Ava256 dataset) and Portrait4Dv2~\cite{deng2024portrait4dv2} as well as the NeRF-based face model HeadNeRF~\cite{hong2022headnerf} in a self-reenactment scenario on persons from the NeRSemble dataset~\cite{kirschstein2023nersemble} Note that the NeRSemble dataset has not been used during training and therefore constitutes an evaluation scenario where both source and driver image are out-of-domain. Ours\textsuperscript{1} is a version of our model that was trained on only 1 input image (see~\cref{sec:data_efficiency}). Our method with 3D lifting (Ours\textsuperscript{3DGAN}) shows better rendering quality than the baselines in these challenging scenarios where input and target view are from opposite sides of the face.}
    \label{fig:4_single_shot_comparison_nersemble}
\end{figure*}

\subsection{Single-shot 3D Head Avatar Creation}
To make Avat3r amenable for inference on only a single input image, we make use of a pre-trained 3D GAN~\cite{yuan2023goae} to first lift the single image to 3D and then render four views of the head. These renderings then constitute the input for Avat3r. We conduct comparisons with the recent 3D-aware portrait animation method GAGAvatar~\cite{chu2024gagavatar}. Specifically, we compare with two version of GAGAvatar: One provided by the authors which is trained on VFHQ~\cite{xie2022vfhq}, and another version, denoted as GAGAvatar\textsuperscript{\textdagger}, that we train on the Ava256 dataset in the same setting as our method. To drive GAGAvatar, we use their monocular FLAME tracker to obtain tracked meshes.
We also compare with another 3D portrait animation method, Portrait4Dv2~\cite{deng2024portrait4dv2}, and HeadNeRF~\cite{hong2022headnerf}.
\cref{fig:4_single_shot_comparison_ava} and \cref{fig:4_single_shot_comparison_nersemble} show qualitative comparisons between our method and the baselines for single input images of hold out persons.
Note that our method performs competitively compared to the single-input baselines despite never being trained for a single-shot scenario. We also include a version of our model that was trained on a single input image without DUSt3R. In general, we find that {\OURS} produces more realistic facial expressions than GAGAvatar and Portrait4Dv2 which are limited by FLAME's expression space. Furthermore, our method allows much more extreme viewpoint changes without sacrificing rendering quality. \cref{tab:single_image_comparison} shows quantitative results. Note that, in contrast to portrait animation methods like GAGAvatar and Portrait4Dv2, our method can benefit when more input views are available (see~\cref{sec:data_efficiency} and tab. 1 in the main paper).

\subsection{Comparison with Monocular Methods}

We compare with the recent monocular approach FlashAvatar and provide it with a full video sequence from NeRSemble. As shown in \cref{fig:iccv_rebuttal_flash_avatar_comparison}, the monocular method performs well from the training view but fails on novel views due to overfitting. Please also refer to the Gaussian Avatar Fusion or HeadGAP papers, which made similar observations. In contrast, our learned reconstruction prior produces a plausible 3D head avatar. Quantitative comparisons are shown in~\cref{tab:single_image_comparison}.

\begin{figure}[htb]
    \centering
    \begin{subfigure}{0.444\linewidth}
        \includegraphics[width=\linewidth]{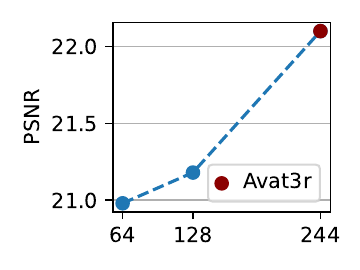}
        \caption{Training subjects}
    \end{subfigure}%
    \begin{subfigure}{0.556\linewidth}
        \includegraphics[width=\linewidth]{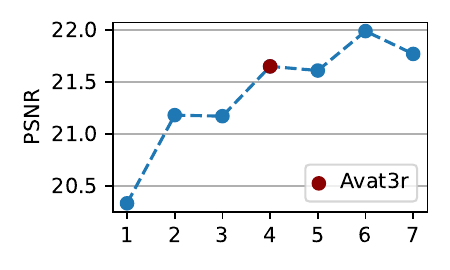}
        \caption{Input views}
    \end{subfigure}
    
    \caption{\textbf{Analysis of Data Efficiency.} We study how reconstruction and animation performance behaves when changing the number of training subjects and input views.}
    \label{fig:4_data_efficiency}
\end{figure}

\begin{figure}[htb]
    \centering
    \includegraphics[width=\linewidth]{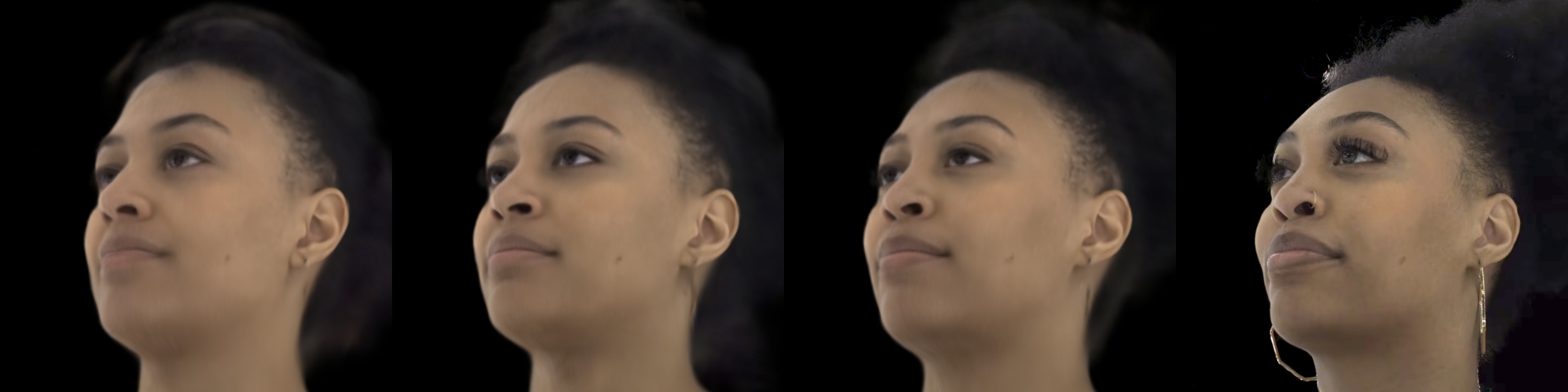}
    \begin{tabularx}{\linewidth}{YYYY}
        64 persons & 128 persons & 244 persons & GT
    \end{tabularx}
    \caption{\textbf{Effect of Number of Train Subjects.} Training on a larger and more diverse set of people enhances the Avat3r's generalization capabilities, as expected. This leads to more accurate reconstructions, with avatars better matching the identities shown in the input images. For instance, when dealing with complex hairstyles, a model trained on a broader range of individuals reproduces the hairstyle more accurately. All ablations are trained without LPIPS loss. }
    \label{fig:x_ablation_num_subjects}
\end{figure}

\begin{figure}[htb]
    \centering
    \includegraphics[width=\linewidth]{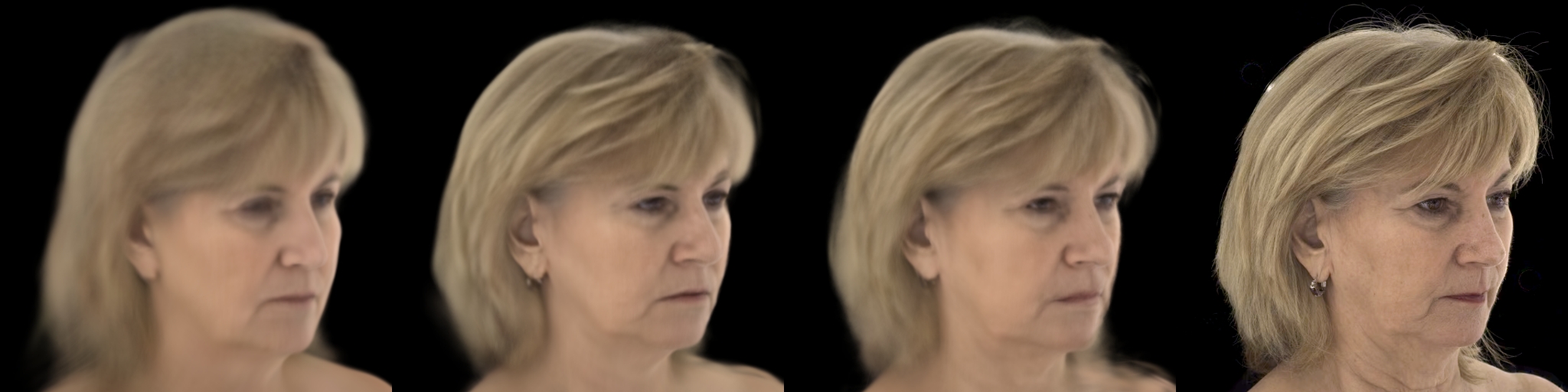}
    \begin{tabularx}{\linewidth}{YYYY}
        1 input image & 4 input images & 7 input images & GT
    \end{tabularx}
    \caption{\textbf{Effect of Number of Input Views.} Training with just a single input image noticeably impairs quality. On the other hand, using more than 4 input images during training does not lead to significant improvements. Models are trained without DUSt3R position maps and without LPIPS loss in the interest of comparability.}
    \label{fig:x_ablation_num_input_views}
\end{figure}

\section{Analysis of Data Efficiency}
\label{sec:data_efficiency}

\subsection{Scaling Subjects and Views from Ava256}
In~\cref{fig:4_data_efficiency} we show how our model scales with the number of training subjects and input views available in the Ava256 dataset. For the analysis on the number of input views, we disable DUSt3R as it produces less reliable position maps for 2 input views, and cannot be executed at all for 1 input view. We see a clear improvement when using more training subjects as well as using more input views. However, further scaling the number of input views also has drawbacks, as it drastically increases runtime due to dense attention inside the transformer and the increased number of Gaussians that have to be rendered. 
We qualitatively analyze the effect of using more train subjects in~\cref{fig:x_ablation_num_subjects} and the effect of the number of input views in~\cref{fig:x_ablation_num_input_views}.

\subsection{Effect of Adding More Neutral Subjects}
\begin{figure}[tb]
    \setlength{\tabcolsep}{0pt}
    \centering
    \begin{tabularx}{\linewidth}{P{0.25\linewidth}P{0.25\linewidth}P{0.25\linewidth}P{0.25\linewidth}}
        \scriptsize{Ours\textsuperscript{3DGAN}} &  \scriptsize{Ours\textsuperscript{3DGAN + 984}} & \scriptsize{Ours\textsuperscript{3DGAN}} &  \scriptsize{Ours\textsuperscript{3DGAN + 984}}\\
    \end{tabularx}
    \includegraphics[width=\linewidth]{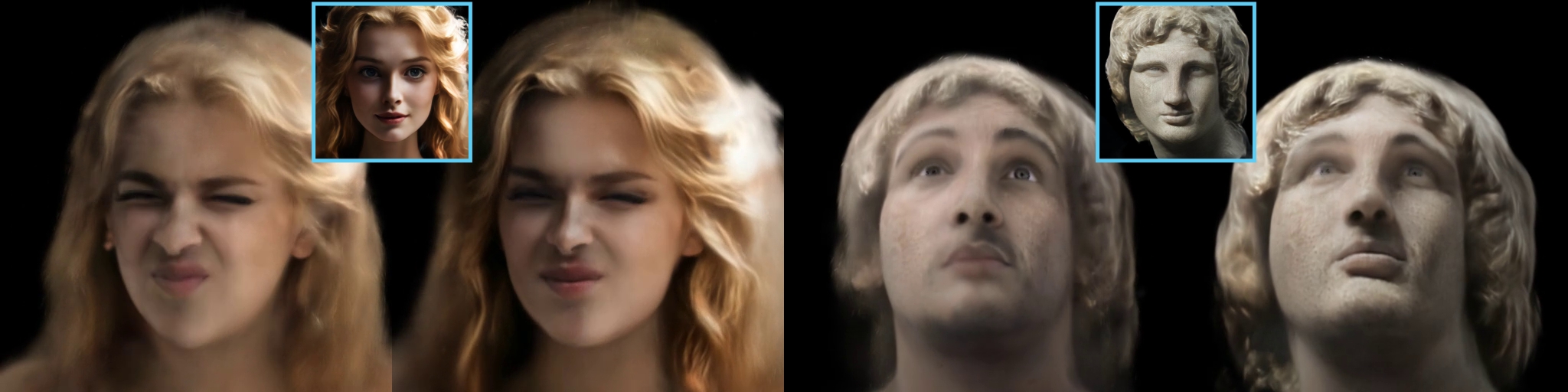}
    \caption{\textbf{Improved identity preservation} by including 984 additional identities with neutral expression during training.}
    \label{fig:iccv_rebuttal_identity_preservation}
\end{figure}

Since Avat3r is only trained on 244 subjects from the Ava256 dataset, it is at risk of overfitting to those identities during training time. In our experiments, we observe that while there is a slight identity shift between the final avatar and the person in the input images, the expression transfer works quite well. We therefore hypothesize that for further improvements on the model's generalization capabilities it is not necessary to also add thousands of expressions for each additional person. To test this, we fine-tune our model with 984 additional identities from an internal dataset — just one expression each — adding only 0.08\% to the training data. As shown in~\cref{fig:iccv_rebuttal_identity_preservation} and confirmed by improved CSIM metrics in~\cref{tab:single_image_comparison}, this small addition noticeably improves identity retention for these challenging cases.

\begin{figure*}
    \centering
    \includegraphics[width=\linewidth]{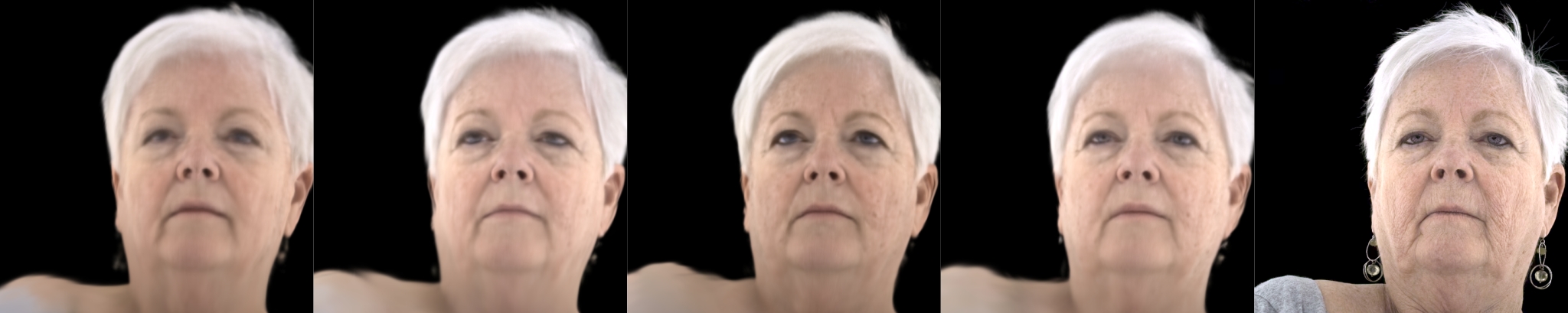}
    \begin{tabularx}{\linewidth}{YYYYY}
        (a) No skip connections & (b) No position skip & (c) No color skip & (d) Ours full & GT
    \end{tabularx}
    \caption{\textbf{Qualitative Ablation of Skip Connections.} Not employing skip connections~(a) causes misalignments, blurry renderings, and a slight color shift. Adding the color skip connection~(b) already noticeably improves sharpness and color fidelity. On the other hand, if the position skip connection is added~(c), geometric details are improved but the overall color slightly off. Using both skip connections~(d) yields the best result.  }
    \label{fig:x_ablation_skip_connections}
\end{figure*}

\begin{table}[htb]
    \centering
    \begin{tabular}{lrr}
        \toprule
            & Creation$\downarrow$ & Driving$\uparrow$ \\
            & in [s]          & in [fps] \\
        \midrule
         HeadNeRF 
            & 65    & \textbf{111}\\
         Portrait4D-v2
            & 0.2   & 4 \\
         GAGAVatar
            & \textbf{0.1}   & 63 \\
         GPAvatar
            & 0.2   & 9.5 \\
         Ours\textsuperscript{1}
            & 1.1   & 53 \\
         Ours\textsuperscript{3DGAN}
            & 17.9  & 7.9\\
         Ours
            & 12.3  & 7.9\\ 
         \bottomrule
    \end{tabular}
    \caption{\textbf{Runtime analysis.} Our method can create a high-quality avatar in a few seconds, and animate it at interactive rates.}
    \label{tab:runtime_analysis}
\end{table}

\section{Inference efficiency \& driving speed.}
While Avatar creation takes several seconds for our method due to DuSt3R and Sapiens, we can cache all activations up to the final cross-attention layers afterwards, leading to expression driving at 7.9fps for our 4-shot model and 53fps for our single-shot model, see~\cref{tab:runtime_analysis}. Runtimes measured on a single RTX3090 GPU.

\renewcommand{\checkbox}{\scalebox{1.5}{$\boxtimes$}}
\renewcommand{\emptybox}{\scalebox{1.5}{$\square$}}

\begin{table}[tb]
    \setlength{\tabcolsep}{2pt}
    \centering
    \resizebox{\linewidth}{!}{
    \begin{tabular}{lccrrrrrr}
        \toprule
             & $\mathcal{C}$ & $\mathcal{P}$
             & \footnotesize{PSNR}$\uparrow$ & \footnotesize{SSIM}$\uparrow$ & \footnotesize{LPIPS}$\downarrow$ & \footnotesize{JOD}$\uparrow$ & \footnotesize{AKD}$\downarrow$ & \footnotesize{CSIM}$\uparrow$
             \\
        \midrule
            No skip & \emptybox & \emptybox
                & 21.39 & 0.740 & 0.456 & 4.99 & 9.24 & 0.60 \\
            No pos. skip & \checkbox & \emptybox
                & 21.76 & 0.746 & 0.443 & 5.03 & 9.04 & 0.611 \\
            No col. skip & \emptybox & \checkbox
                & 21.55 & 0.745 & 0.435 & 5.00 & \textbf{7.69} & 0.648 \\
            Avat3r & \checkbox & \checkbox
                & \textbf{22.05} & \textbf{0.751} & \textbf{0.421} & \textbf{5.15} & 7.99 & \textbf{0.689}\\ %
        \bottomrule
    \end{tabular}
    }
    \caption{\textbf{Quantitative Ablation of Skip Connections.} We analyze the effect of the color~($\mathcal{C}$) and position~($\mathcal{P}$) skip connections. All ablation models are trained without LPIPS loss. Metrics are computed on $667 \times 667$ renderings.}
    \label{tab:x_skip_connection_ablations}
\end{table}

\begin{figure*}[htb]
    \centering
    \includegraphics[width=\textwidth]{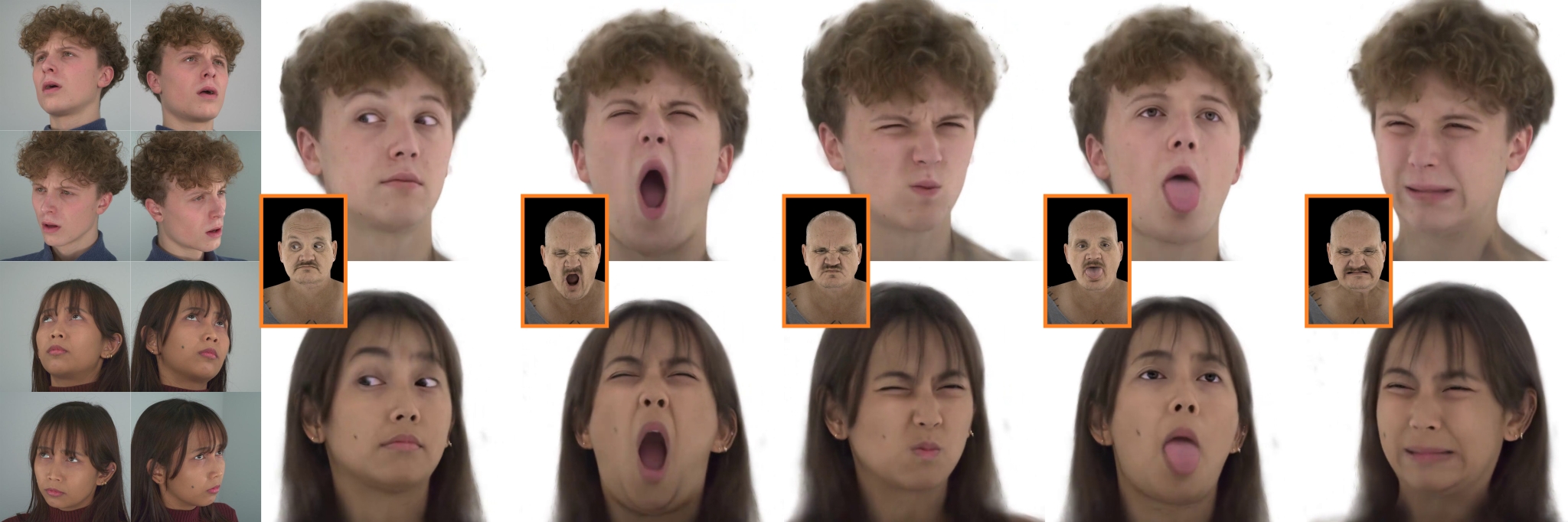}
    \begin{tabularx}{\linewidth}{P{0.15\linewidth}P{0.78\linewidth}}
Input Images & Animated 3D Head Avatars
\end{tabularx}
    \vspace{-0.5cm}
    \caption{\textbf{Performance on NeRSemble dataset.} We show reconstructed and animated avatars using 4 images from the NeRSemble dataset. Note that this dataset was not used during training and contains images with different lighting conditions, viewpoints, and camera intrinsics than the Ava-256 dataset that was used to train {\OURS}.}
    \label{fig:04_nersemble_results}
    \vspace{-0.5cm}
\end{figure*}

\section{Effect of Skip Connections}

We analyze the effect of the proposed skip connections, i.e., omitting Eq.~(11), Eq.~(12), or both of the main paper. The results are listed in~\cref{tab:x_skip_connection_ablations}. We observe a noticeable hit in performance when either skip connection is removed. Furthermore, we qualitatively analyze the effect of skip connections in~\cref{fig:x_ablation_skip_connections}.

\section{Training Details}

\paragraph{Dataset processing.}
We use the 4TB version of the Ava256 dataset~\cite{ava256} which contains 256 persons, 80 cameras, and roughly 5000 frames per person that are sampled at 7.5 fps. We compute foreground segmentation masks with BackgroundMattingV2~\cite{BGMv2} and replace the background in all images with black pixels. We use the provided tracked mesh to find a $512 \times 512$ head-centered square crop for input images and $667 \times 667$ head-centered square crop for supervision views. This ensures that the pixels in the input images are used efficiently to show as much as possible of the head, leading to more 3D Gaussians. The reason for also cropping the target images is to remove parts of the torso, as it is not the focus of this work.
\paragraph{DUSt3R and Sapiens.}
Since both DUSt3R~\cite{wang2024dust3r} and Sapiens~\cite{khirodkar2025sapiens} are expensive foundation models, we pre-compute the position and feature maps for the input frames.
For Dust3r, it is prohibitive to pre-compute all possible combinations of 4 input views out of the available 80 cameras. Instead, we choose 3 "reasonable partner views" for each input and only store the position map for that viewpoint. This assigns each input view exactly one position map, which is conceptually wrong since the position map from DUSt3R should depend on the other 3 selected views. Nevertheless, we did not observe any disadvantages from this simplification strategy.

\begin{table}[tb]
    \centering
    \resizebox{\linewidth}{!}{
    \begin{tabular}{lll}
        \toprule
        & Hyperparameter & Value \\
        \midrule
            \multirow{5}{*}{\rotatebox[origin=c]{90}{\parbox[c]{2cm}{\centering \small ViT \& GRM}}} 
                & ViT patch size & $8 \times 8$ \\
                & hidden dimension $D$ & 768 \\
                & \#self-attention layers & 8 \\
                & \#cross-attention layers & 8 \\
                & \#GRM transformer upsampler step & 1 \\
         \midrule
            \multirow{5}{*}{\rotatebox[origin=c]{90}{\parbox[c]{2cm}{\centering \small Input \& Output}}} 
                & Sapiens version & 2b \\
                & Sapiens feature dimension & 1920 \\
                & Input image resolution & $512 \times 512$ \\
                & Gaussian attribute map resolution & $512 \times 512$ \\
                & Train render resolution & $667 \times 667$ \\

         \midrule
            \multirow{3}{*}{\rotatebox[origin=c]{90}{\parbox[c]{1.5cm}{\centering \small Expression MLP}}} 
                & Dimension of expression code & 256 \\
                & \#expression sequence MLP layers & 2 \\
                & Dimension of expression sequence MLP & 256 \\
                & Expression sequence MLP activation & ReLU \\
         \bottomrule
         
    \end{tabular}
    }
    \vspace{-0.3cm}
    \caption{\textbf{Hyperparameters.}}
    \vspace{-0.5cm}
    \label{tab:x_hyperparameters}
\end{table}

\paragraph{Head-centric coordinates.}
We further simplify the task by factoring the head poses from the provided tracked mesh into the camera poses instead of letting the network predict them. That way, our model can always predict the head in canonical pose, making the task easier. This is possible because modeling the torso, which in head-centric coordinates moves a lot when the person shakes their head, is not the focus of this work. 
\paragraph{Expression codes.}
Our architecture is agnostic to the specific choice of animation signal. For experiments on ava256, we used the dataset's expression codes that were originally predicted by a generalized expression encoder providing a driving signal beyond FLAME's topology. For experiments on NeRSemble and in-the-wild driving videos, we fine-tuned our model using FLAME codes obtained by running GAGAvatar's version of Metrical Tracker~\cite{MICA:ECCV2022, chu2024gagavatar}. This shows that Avat3r learns a general notion of facial expressions that can be adapted to fit a specific driving signal.

\paragraph{k-farthest viewpoint sampling.}
To ensure that the 4 input images always follow a reasonable viewpoint distribution, we employ k-farthest viewpoint sampling. Specifically, we first start from a random camera and collect a set of 10 candidate cameras that are evenly spread out using farthest point sampling. From this candidate set, we then randomly select 4 cameras as input. This two-stage approach ensures that the input cameras are sufficiently random during training but also reasonably spread out to avoid seeing a person only from one side. During sampling input viewpoints, we exclude cameras that only observe the person from the back since those are not realistic inputs during test-time.
\paragraph{Input timestep sampling.}
To improve robustness of our model, we sample different timesteps for each of the 4 input images. This ensures that the model can deal with inconsistencies in the input. To maximize the diversity in the input expressions, we uniformly sample 10 timesteps in the segments:
\verb+EXP_eye_wide+, \verb+EXP_tongue001+, and \verb+EXP_jaw003+ from the recordings of the Ava256 dataset. This covers the most extreme facial expressions while avoiding having to pre-compute DUSt3R and Sapiens maps for every single image in the dataset.

\paragraph{Speed.}
To speed-up training, we employ the 3D Gaussian Splatting performance improvements of DISTWAR~\cite{durvasula2023distwar}. \\

\section{Hyperparameters}

In~\cref{tab:x_hyperparameters}, we list the most important hyperparameters for training Avat3r.

\end{document}